\documentclass{article}
\usepackage{ijcai24}

\usepackage{times}
\usepackage{soul}
\usepackage{url}
\usepackage[hidelinks]{hyperref}
\usepackage[utf8]{inputenc}
\usepackage[small]{caption}
\usepackage{graphicx}
\usepackage{amsmath}
\usepackage{amsthm}
\usepackage{booktabs}
\usepackage{multirow}
\usepackage{algorithm}
\usepackage[switch]{lineno}
\usepackage{subcaption}
\usepackage{algorithmicx}
\usepackage{amsmath}
\algnewcommand{\LeftComment}[1]{\Statex \(\triangleright\) #1}
\usepackage[noend]{algpseudocode}
\usepackage{comment}

\newcommand{\etal}{{\textit{et al.}}}

\title{Minimally Supervised Learning using Topological \\Projections in Self-Organizing Maps}

\author{
Zimeng Lyu\and
Alexander Ororbia\and
Rui Li\and
Travis Desell\\
\affiliations
Rochester Institute of Technology\\
\emails
zimenglyu@mail.rit.edu,
agovcs@rit.edu,
rxlics@rit.edu,
tjdvse@rit.edu
}

\begin{document}

\maketitle

\begin{abstract}
Parameter prediction is essential for many applications, facilitating insightful interpretation and decision-making. However, in many real life domains, such as power systems, medicine, and engineering, it can be very expensive to acquire ground truth labels for certain datasets as they may require extensive and expensive laboratory testing. In this work, we introduce a semi-supervised learning approach based on topological projections in self-organizing maps (SOMs), which significantly reduces the required number of labeled data points to perform parameter prediction, effectively exploiting information contained in large unlabeled datasets. Our proposed method first trains SOMs on unlabeled data and then a minimal number of available labeled data points are assigned to key best matching units (BMU). The values estimated for newly-encountered data points are computed utilizing the average of the $n$ closest labeled data points in the SOM's U-matrix in tandem with a topological shortest path distance calculation scheme. Our results indicate that the proposed minimally supervised model significantly outperforms traditional regression techniques, including linear and polynomial regression, Gaussian process regression, K-nearest neighbors, as well as deep neural network models and related clustering schemes.

\end{abstract}

\section{Introduction}
\label{sec:intro}

Parameter prediction is a fundamental aspect of numerous data science applications, enabling the interpretation of insights as well as aiding in decision-making processes~\cite{barnes2016real}\cite{candanedo2017data}\cite{jiang2021deep}. In the financial sector, it is crucial for predicting future stock prices or investment returns~\cite{thakkar2021fusion}. It is also instrumental in risk management and insurance, where projecting potential losses is essential when designing robust insurance policies and mitigation strategies~\cite{kure2022asset}. A specific application of parameter estimation is often seen in power industries, particularly in operations involving coal-fired power plants~\cite{smrekar2010prediction}\cite{adams2020prediction}. These facilities utilize parameter estimation to estimate future demand, forecast fuel properties, anticipate maintenance needs, and assess potential environmental impacts. By predicting these values, plants can optimize operational efficiency, plan budgets, and better comply with environmental regulations. In effect, parameter estimation translates data-driven insights into quantifiable future predictions, thus providing organizations with the necessary tools to make informed and forward-looking decisions.

However, in practice, acquiring labeled data can be extremely expensive and time-consuming. Serving as the motivating example and data generating process for this study, a coal power plant needs to analyze coal property decompositions from a sensor which provides $512$ spectra readings per minute as coal flows through it in order to make real-time operational decisions. This power plant has $12$ cyclones, of which only two were installed with expensive sensors that record time series of these $512$ channels of spectra readings. From the time that the sensors were installed in 2020, only three field tests were conducted to collect samples that were sent to a lab for analysis. Concretely, these required human specialists to travel to the power plant in order to find and collect representative coal samples and then send those samples to a lab for analysis; the lab analyzed the coal samples and sent the results back after a few weeks. Each field test was only able to acquire around $20$ coal property analysis data points, in each field trip for each cyclone. 


\begin{figure*}[htbp]
  \centering
  \includegraphics[width=0.685\textwidth]{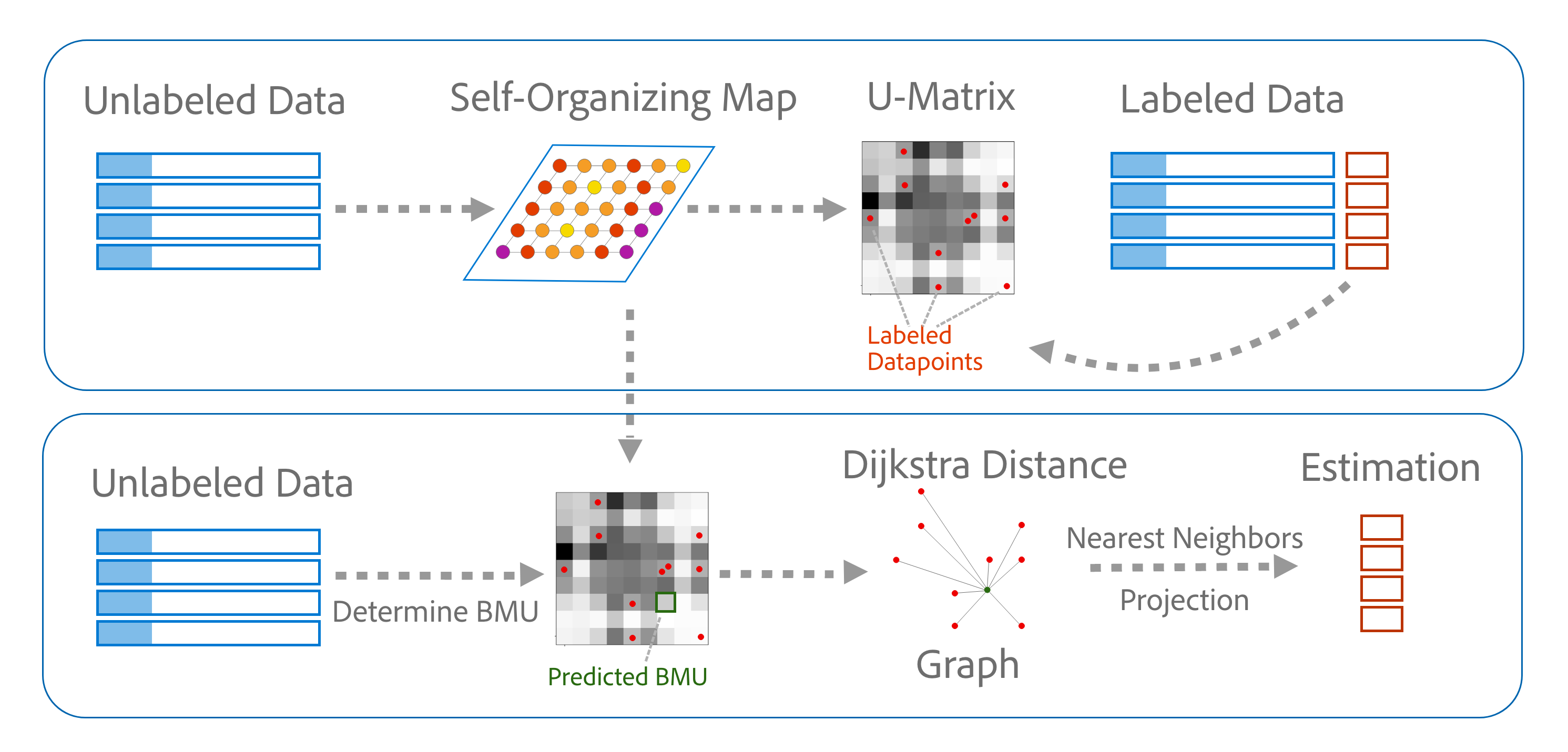}
  \vspace{-0.2cm}
  \caption{The proposed minimally supervised SOM modeling framework.}
  \label{fig:som}
  \vspace{-0.3cm}
\end{figure*}

Self-organizing maps (SOMs)~\cite{kohonen1982self} are a type of artificial neural network that learn in an unsupervised fashion, creating a low-dimensional representation of input data, usually within a two-dimensional space. SOMs are well-known for their unique ability to maintain the topological and metric relationships of the original high-dimensional data, making them a powerful tool for visualizing and interpreting complex datasets, particularly those with many attributes or observed variables~\cite{kohonen2013essentials}\cite{vesanto2000clustering}\cite{kohonen1996engineering}.

The main hypothesis of this work is that a SOM can be trained using a large amount of unlabeled data, learning how properly represent the topology of the input data space, and be readily used for cases where labeled data is limited. This has resulted our design of a novel approach for semi-supervised learning scheme -- one that we also call minimally-supervised -- that utilizes topological projections within the SOM framework. For the coal data example, only 67 labeled data points were available for a dataset of over 20k samples. In this algorithm, the SOM is first trained on unlabeled data for clustering and then the few available labeled data points are mapped to the trained SOM's topology. Finally, we make use of the distance relationship between an unknown data point's best matching unit (BMU) and its closest $N$ labeled data points' BMUs to dynamically craft parameter value predictions. The performance of the proposed method is empirically tested on coal fired power plant spectra data as well as appliance energy consumption data; our results demonstrate that our minimally-supervised learning approach works better other powerful supervised and unsupervised learning methods.


\begin{algorithm*}[!t]
    \begin{algorithmic}[1]
        \caption{SOM Topological Projection}
        \label{alg:som}
        \footnotesize
        \Function{SOM}{$X\textrm{-}{unlabeled}$, $X\textrm{-}{labeled}$, $Y\textrm{-}{labeled}$}
            \State $som = SOMClustering.fit(X\textrm{-}{unlabeled})$
            \Comment{\emph{Cluster unlabeled data with SOM}}
            \State $U\textrm{-}Matrix = som.getU\textrm{-}Matrix()$
            
            \State $DistanceGraph$ = $DijkstraGraph(UMatrix)$
            \Comment{\emph{Build Dijkstra Shortest Paths Distance Graph}}
            
            \State $PairwiseDistance = DistanceGraph.getPairwiseDistance()$
            \Comment{\emph{Shortest distance between any two nodes}}
            
            \State $Labeled\textrm{-}BMUs = som.transform(X\textrm{-}{labeled})$
            \Comment{\emph{Fit labeled data on trained SOM}}
            
            \State $ClosestNeighbors = findClosestNeighbor(PairwiseDistance, Labeled\textrm{-}BMUs, N)$
            
            \Comment{\emph{Find N closest neighbors and their pairwise distances for each node}}

            \State $EstimatedValues = makeTable(ClosestNeighbors, Y\textrm{-}{labeled}, N)$
            \Comment{\emph{Calculate estimated values for each node}}

            \For{$x$ in $X\textrm{-}{unlabeled}$}
                \State $x\textrm{-}BMU = som.transform(x)$
                \Comment{Get unknown data point's BMU}
                \State $y\textrm{-}{estimated} = EstimatedValues[x\textrm{-}BMU]$
                \Comment{Look up its estimated values}
            \EndFor
        \EndFunction

        \Function{MakeWeightedNeighborTable}{$ClosestNeighbors$, $Y\textrm{-}{labeled}$, $N$}
            \For{$x$ in $X\textrm{-}{unlabeled}$}
                \For{$n$ in $N$}
                    \State $weight = 1/ClosestNeighbors[n]$
                    \Comment{\emph{Weight is the inverse of pairwise distance}}
                    \State $distance = Y\textrm{-}{labeled}[n]$
                    \State $SumWeight += weight$
                    \State $SumDistance += distance$
                \EndFor
                \State $Estimation[x] = SumDistance / SumWeight$
                \Comment{Get weighted average of closest neighbors}
            \EndFor
            \State $return Estimation$
        \EndFunction

        \Function{MakeLineSearchTable}{$ClosestNeighbors$, $Y\textrm{-}{labeled}$, $LineSearchMethod$}
            \For{$x$ in $X\textrm{-}{unlabeled}$}
                \If{$LineSearchMethod == Linear$}
                    \State $model = LinearRegressor$
                \EndIf
                \If{$LineSearchMethod == Polynomial$}
                    \State $model = PolynomialRegressor$
                \EndIf
                \State $model.fit($ClosestNeighbors$, $Y\textrm{-}{labeled}$)$
                \State $Estimation[x] = model.predict[0]$
                \Comment{The estimated value is when x=0 in regression model}
            \EndFor
            \State $return Estimation$
        \EndFunction
    \end{algorithmic}
\end{algorithm*}

\section{Related Work}
\label{sec:related_work}

Semi-supervised learning / minimally supervised learning primarily deals with the problem of learning from a limited number of labeled samples. It is widely used for classifications related tasks, such as natural language processing (NLP)~\cite{zhang2021minimally}\cite{qiang2023minimally}, image and video analysis~\cite{wang2020focalmix}\cite{wang2020semi}, and anomaly detection~\cite{villa2021semi}\cite{jiang2021semi}\cite{azzalini2021minimally}. For regression problems, semi-supervised learning is mostly used for computer vision related tasks. Dai~\etal~propose Uncertainty-Consistent Variational Model Ensembling using Beyesian networks to improve uncertainty estimations~\cite{dai2023semi}. Rezagholiradeh~\etal~use semi-supervised generative adversarial networks for regression~\cite{rezagholiradeh2018reg}. 
Very few related works use semi-supervised regression for data science. A.H.de Souza Júnior~\etal~porpose minimal learning machine for supervised distance-based non-linear regression and classification on multidimensional response spaces~\cite{de2015minimal}.
Levati{\'c}~\etal~use semi-supervised multi-target regression (MTR) for multi-target regression~\cite{levatic2015semi}. Kostopoulos~\etal~use semi-supervised learing to predict the final grade of students who take online courses~\cite{kostopoulos2019semi}. Meattini~\etal~proposed minimally supervised regression model for robot hand grasping control~\cite{meattini2022semg}.
With respect to the SOM, or Kohonen map, the original efforts behind its proposal and initial construction focused on unsupervised clustering~\cite{kohonen1982self}. However, later work showed that the SOM could be adapted for semi-supervised learning, combining it with a K-nearest neighbors (KNN) classifier to tackle different categorization tasks~\cite{silva2011som}\cite{tian2014anomaly}\cite{suchenwirth2014large}. For regression problems, Hsu~\etal~proposed a two-stage architecture using an SOM and support vector regressor for stock price prediction~\cite{hsu2009two}. Riese~\etal~proposed a supervised and semi-supervised SOM algorithm that was capable of unsupervised, supervised, and semi-supervised classification as well as regression on high-dimensional data~\cite{riese2019supervised}. The SOM has also be seen integration/fusion with other regression methods, such as support vector regression~\cite{che2012adaptive}\cite{dong2015novel} and geographically-weighted regression~\cite{wang2020elucidating} for domain-specific applications.  To our knowledge, this is the first work to utilize the SOM topology itself to perform parameter prediction, particularly in the context of minimally available annotated target samples.
\section{SOM Topological Projections}
\label{sec:topo_projections}

A SOM is a neural system that learns/adapts in an unsupervised fashion; it is a model that is used primarily for visualization and dimensionality reduction~\cite{kohonen1990self}. SOMs are able to transform high-dimensional data samples into a low-dimensional map (usually two-dimensional) which can be represented visually through what is known as a U-Matrix -- the learned map preserves the topological and metric relationships of the original data points. This makes SOMs particularly useful for visualizing complex datasets and recognizing clusters or patterns within them. 

In Kohonen maps, units are typically arranged using a square (each unit has four neighbors) or a hexagonal (each unit has six neighbors) map. As the SOM is trained, samples are matched to their BMU and the BMU and its neighboring units within a radius are moved towards the sample; the neighbors are pulled to a lesser degree depending on how far away from the BMU they are within the map. This results in neighboring units typically being closer to each other in the map while still allowing the overall map to represent the topology of the sample space.

\begin{figure}[!t]
  \centering
  \includegraphics[width=0.5\columnwidth]{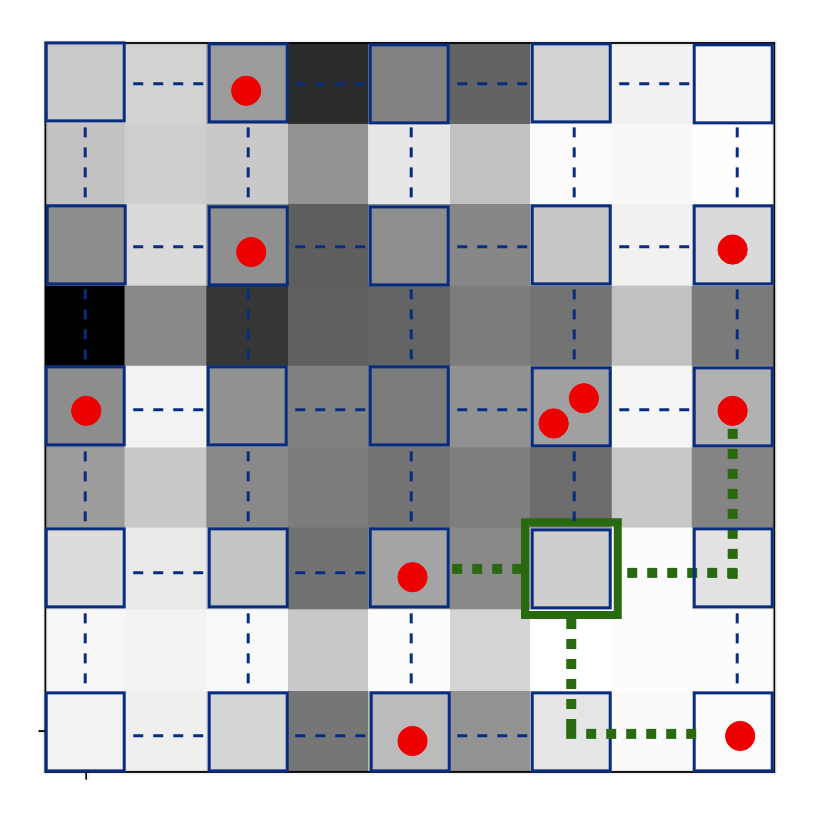}
  \caption{An example topological projection of an unlabeled data point to a trained SOM topology with mapped labeled data.}
  \label{fig:umatrix}
  \vspace{-0.3cm}
\end{figure}

This work utilizes the map topology to project values for unlabeled data patterns given the integration of a small number of labeled data points assigned to the map. Figure~\ref{fig:som}~presents a flowchart of our proposed method and Algorithm~\ref{alg:som} formally depicts the methodology. Concretely, the SOM is first trained using unlabeled data and the labeled data points are mapped to their best matching units wtihin the SOM (marked as red dots in the U-Matrix depicted in Figure \ref{fig:umatrix}). Parameters for unlabeled data points during inference can then be estimated by projecting values from their nearest neighbors in the SOM as shown in Figure~\ref{fig:umatrix}. The squares with blue borders are the SOM units and the squares between SOM nodes represent the distance between units, visualized in varying shades of gray; darker colors represent a greater measured distance between the nodes on either side of the cell. The red dots represent the small set of labeled datapoints that are fitted into the trained SOM.

Following this, also shown in Figure \ref{fig:umatrix}, an example new unlabeled data point A is matched to its best matching unit (the green cell). Dijkstra's algorithm is then used to calculate (given the distance between neighboring units in the SOM graph) the shortest paths' pairwise distance between the green node and all of the nodes that contain labeled data (note that this step can be calculated once for each unit in the SOM and have the distances values saved to significantly improve inference performance). The $N$ nearest neighbors -- in this case, $N=3$ -- are selected (paths shown by green dotted lines) and then used for final parameter prediction.

This method is flexible given that any distance metric could be employed in order to calculate distance values in the topological projection; in this work, we use the Euclidean distance as it provided best results.  Furthermore, the topological projection utilized to predict parameter values can also be carried out in different ways. This work investigates using a weighted average of the nearest neighbors as well as a regression based on nearest neighbor distance measurements.

\subsection{Weighted Average Projection}
\label{sec:weighted_average_projection}

Given the $N$ nearest labeled neighbors, with parameters $n_p$, a weighted average estimation for each parameter $e_p$ can be predicted given the distance from each unlabeled data point's BMU to each of that BMU's neighbors, $d(BMU, n)$:
\[
    e_p= \frac{\sum_{n=1}^{N} n_p\cdot{\frac{1}{d(BMU, n)}}}
                    {\sum_{n=1}^{N}{\frac{1}{{d(BMU, n)}}}}.
\]

\subsection{Regression Projection}
\label{sec:regression_projection}

Alternately, each neighbor distance and labeled parameter value can be used as x/y pairs to train regression estimators. If the x value is the distance between the unknown node and its neighbors, then $f(x)$ is the parameter value, which is known based on the neighbors. This facilitates the use of regression in the context of conducting parameter prediction, where the the estimated value is $f(x=0)$.  This work investigates estimating parameters for both linear and polynomial regressors.

\section{Experimental Setup}
\label{sec:experiments}

\subsection{Dataset}
\label{sec:data}
This work utilizes two real-world datasets from the energy domain. The first one was a primary motivation for this work and was collected from a coal fired power plant's cyclones. The unlabeled input data is per-minute coal sample spectra readings, each with $512$ channels, which was collected from the years 2021 through 2023. This unlabeled data consists of more than $20$K data points. Of these, only $67$ data points were taken as samples and sent to a lab for coal property analysis in order to generate labeled values for training a predictor. The $13$ labeled coal properties for prediction provided by the laboratory analysis were: \emph{BaseAcidRatio, AshContent, BTU, H2OContent, NaContent, FeContent, AlContent, CaContent, Kcontent, MgContent, SiContent, SOContent, TiContent}.

The second dataset is appliance energy usage in a low energy house~\cite{candanedo2017data}. The appliance energy usage can be affected by things such as the temperature and humidity of the environment in the low energy house. The energy dataset was recorded every $10$ minutes and also has approximately $20$K data points. The dataset contains $27$ input parameters, including energy use of light fixtures, the humidity and temperature of different rooms in the house, as well as weather data from outside of the house. This dataset serves as a baseline/benchmark given that all data points contain our prediction parameter target (appliance energy usage), which lets us remove this parameter target from varying amounts of data points, allowing us to examine how well our algorithm performs with more or less data (given its semi-supervised nature), as compared to other benchmark methods.

\subsection{Experimental Setting and Preparation}
\label{sec:exp_setup}
This work explored using both regression (supervised) methods as well as unsupervised clustering methods. For the supervised regression tasks, the labeled data ($67$ datapoints for the coal data and varying quantities for the appliance data) were divided into training and test datasets. The training consisted of $80$\% of entire labeled dataset ($53$ samples for the coal data), and each test dataset had the remaining $20$\% of the labeled dataset ($14$ datapoints for the coal data). The supervised methods compared against included linear regression, polynomial regression, a deep neural network (DNN) regressor, Gaussian process regression (GPR), and K-Nearest neighbors (KNN). For unsupervised learning, we compare the clustering performance between Density-Based Spatial Clustering of Applications with Noise (DBSCAN) and SOM. Unfortunately most of the semi-supervised learning models are designed for computer vision or NLP applications, and the code for some semi-supervised learning work for data science was not available~\cite{de2015minimal}\cite{li2017learning} for us to compare to. 

Each experiment was repeated $10$ times (each experimental trial seeded uniquely) on a MacBook Pro with Intel Core i9 processors. Code is available at [REMOVED for REVIEW]. 
We utilized $k$-fold ($k=5$) cross validation for hyperparameter tuning as well as to select the optimal model configuration. The  best model chosen was then tested on the testing dataset. The input dataset is high-dimensional for the coal data ($512$ parameters); therefore, all of those regression tasks used the top principal components (PCs) which represented $80$\% of the variance from principle components analysis (PCA) used to conduct dimensionality reduction. For the coal data, the top $31$ PCs represented $80$\% of the data's total variance. 

As data normalization methods can affect model training results given that they might alter the original data distribution, all experiments were performed using min-max scaling (or feature scaling), standardization (z-score normalization), and robust normalization.  Min-max scaling does not change the data distribution, however it can be heavily affected by large outliers. Standard normalization assumes that the feature distribution is approximately Gaussian and is less sensitive to outliers as compared to min-max scaling. Robust normalization utilizes the median and interquartile range to transform the data and is the most robust to data outliers out of the three normalization  schemes. 

\section{Results}
\label{sec:results}
\subsection{SOM Hyperparameters}
Apart from the SOM training hyperparameters (e.g., learning rate, radius, training epochs etc), which can be tuned in experiments, we found that two major hyperparameters significantly affected the performance of our proposed method. 

\begin{figure}[!t]
  \centering
  \begin{subfigure}[b]{0.35\textwidth}
    \centering
     \includegraphics[width=0.935\textwidth]{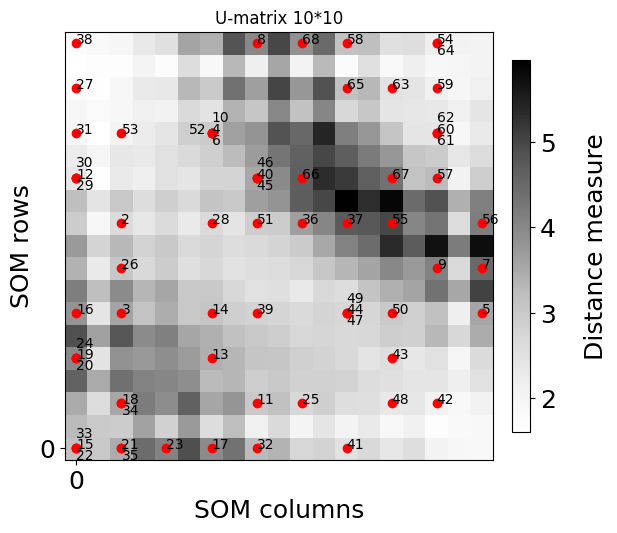}
    \caption{U-Matrix size 10x10}
    \label{subfig:umatrix_15}
  \end{subfigure}
  
  \begin{subfigure}[b]{0.37\textwidth}
    \centering
    \includegraphics[width=0.935\textwidth]{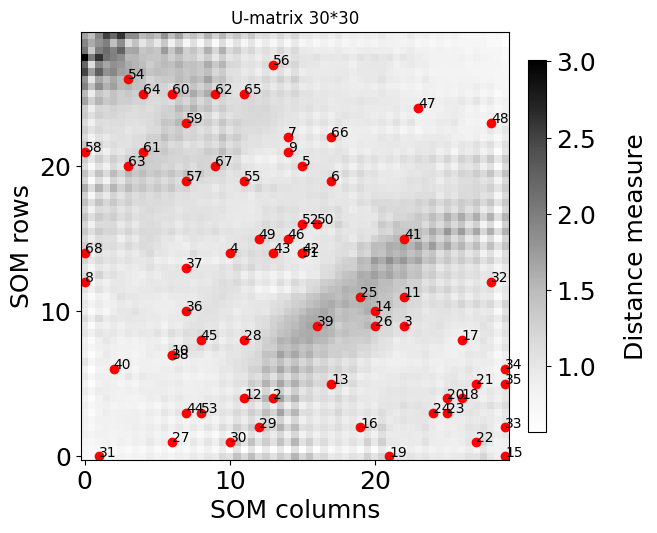}
    \caption{U-Matrix size 30x30}
    \label{subfig:umatrix_30}
  \end{subfigure}
  \vspace{-0.25cm}
  \caption{Labeled data point mappings in SOMs trained on the unlabeled coal data with different grid sizes.}
  \label{fig:two_umatrix}
  \vspace{-0.3cm}
\end{figure}

The first parameter (that our scheme is sensitive to) is the SOM's grid size, as we found that this significantly affects the clustering results  as well as how spread out the labeled data points are. Figure~\ref{fig:two_umatrix} presents how the same labeled data points are distributed on SOMs with size $10\times10$ and $30\times30$ (from the perspective of the coal data set). The two SOMs are trained with the same training hyper parameters and, in the figure, the red dots mark labeled data points. The numbers next to the red dots are the identifier (ID) of each labeled data point. In the $10\times10$ SOM U-Matrix, many labeled data points are clustered together, such as data point $ID=4$, $6$, $10$ and $52$, where data points are more spread out in the $30\times30$ SOM U-Matrix. The more spread out the labeled data points are, the better the parameter value estimation result is, since the SOM more accurately reflects the sample space topology. 

The second parameter our method is sensitive to is the number of neighbors used for weighted average parameter value estimation. Table~\ref{tab:som_rmse}  further depicts that increasing the number of neighbors used in the topological projection does not necessarily yield better value estimation results. Our results indicate that it is more ideal to use the labeled data points near the unknown point for the weighted average estimation. 

Table~\ref{tab:som_rmse} shows the average prediction root mean squared error (RMSE) using different SOM sizes and number of nearest neighbors over $10$ repeated runs using different data normalization methods for the coal dataset. The best RMSE for each of the different data normalization methods are marked in bold font. Different data normalization schemes affect the final optimal SOM model due to their impact on the outliers as well as distribution of the original data~\cite{singh2020investigating}. Nevertheless, the best RMSE for using three normalization methods are fairly close; however, using min-max scaling gives the best results.

\begin{table}[!t]
\centering
\scriptsize
\caption{SOM validation RMSE under various data normalizers, topologies, and number of nearest neighbors $N$ used for estimation.}
\label{tab:som_rmse}
\begin{tabular}{c|*{6}{c}}
\hline
    & N=3 & N=5 & N=7 & N=10 & N=12 & N=15 \\
\hline
\multicolumn{7}{c}{Minmax Normalization} \\
\hline
10x10 & 7.64e2  & 7.28e2  & 7.09e2  & 7.29e2  & 7.38e2  & 7.48e2 \\
15x15 & 8.71e2  & 8.20e2  & 8.33e2  & 8.33e2  & 8.31e2  & 8.06e2 \\
20x20 & 8.37e2  & 7.87e2  & 7.31e2  & 7.29e2  & 7.28e2  & 7.31e2 \\
25x25 & 6.50e2  & 6.58e2  & \bf{6.44e2}  & 6.69e2  & 6.79e2  & 6.65e2 \\
30x30 & 8.79e2  & 7.56e2  & 7.97e2  & 8.61e2  & 8.61e2  & 8.58e2 \\
\hline
\multicolumn{7}{c}{Standard Normalization} \\
\hline
10x10 & 8.14e2  & 7.61e2  & 7.77e2  & 7.74e2  & 7.69e2  & 7.81e2 \\
15x15 & 7.71e2  & 7.43e2  & 7.59e2  & 7.70e2  & 7.78e2  & 8.15e2 \\
20x20 & 8.07e2  & 7.85e2  & 7.50e2  & 7.57e2  & 7.72e2  & 7.64e2 \\
25x25 & 7.05e2  & 7.25e2  & 7.23e2  & 7.11e2  & 7.43e2  & 7.27e2 \\
30x30 & 6.68e2  & 7.87e2  & 7.43e2  & \bf{6.66e2}  & 6.76e2  & 6.96e2 \\
\hline
\multicolumn{7}{c}{Robust Normalization} \\
\hline
10x10 & 7.76e2  & 7.67e2  & 7.93e2  & 7.88e2  & 7.77e2  & 7.86e2 \\
15x15 & 8.06e2  & 7.92e2  & 8.04e2  & 8.00e2  & 7.98e2  & 7.96e2 \\
20x20 & 7.08e2  & \bf{6.83e2}  & 6.91e2  & 7.17e2  & 7.27e2  & 7.36e2 \\
25x25 & 6.90e2  & 7.04e2  & 7.11e2  & 7.42e2  & 7.30e2  & 7.44e2 \\
30x30 & 9.74e2  & 8.77e2  & 8.15e2  & 8.76e2  & 8.59e2  & 8.57e2 \\
\hline
\end{tabular}
\end{table}

\subsection{Coal Dataset Results}
Weighted averaging (WAVG) as well as linear and polynomial regression were tested as our scheme's topological projection on the coal dataset. The experimental SOM sizes and number of nearest neighbors tested were the same as those in Table \ref{tab:som_rmse}. Table \ref{tab:som_wavg_ls} shows the best RMSE for all experiments using weighted neighbors and both regression methods\footnote{Full result tables are included in the Supplementary material.}. The weighted neighbor approach performs the best as compared to linear and polynomial regression under all three data normalization methods. Therefore, we used this scheme for the remaining experiments.

\begin{table}[!t]
\centering
\scriptsize
\caption{SOM weighted average versus line search.}
\label{tab:som_wavg_ls}
\vspace{-0.2cm}
\begin{tabular}{r|*{3}{c}}
\hline
    & Minmax & Standard & Robust \\
\hline
WAVG & 6.44e2 & 6.66e2 & 6.83e2 \\
Linear & 7.75e2 & 7.62e2 & 7.90e2 \\
Polynomial & 1.77e3 & 1.48e3 & 1.65e3 \\
\hline
\end{tabular}
\end{table}

\begin{table}[!t]
\centering
\scriptsize
\caption{Random guessing baseline RMSE.}
\label{tab:random_guess}
\vspace{-0.2cm}
\begin{tabular}{c|*{2}{c}||c|*{2}{c}}
\hline
    & $X{\sim}U$ & $X{\sim}N$ & & $X{\sim}U$ & $X{\sim}N$ \\
    \hline
    B/A & 0.05 & 0.03 & Ca & 7.58e5 & 7.91e5 \\
    Ash & 1.17 & 0.68 & K & 1.01e5 & 1.25e5 \\
    BTU & 9.26e4 & 6.22e4  & Mg & 1.76e5 & 1.02e5 \\
    H2O & 7.22 & 3.62  & Si & 4.91e7 & 2.63e7 \\
    Na & 1.27e6 & 1.11e6 & SO & 9.39e6 & 5.72e6 \\
    Fe & 3.55e6 & 1.79e6 & Ti & 9.85e3 & 8.90e3 \\ \cline{4-6}
    Al & 3.74e6 & 3.11e6 & Average & 5.25e6 & 3.01e6 \\

\hline
\end{tabular}
\vspace{-0.2cm}
\end{table}

\begin{table*}[!t]
\centering
\scriptsize
\caption{Min-max RMSE results.}
\label{tab:minmax_rmse}
\vspace{-0.2cm}
\begin{tabular}{c|*{10}{|c}}
\hline
& \multirow{2}{*}{Linear} & \multirow{2}{*}{Polynomial} & \multicolumn{2}{c|}{GPR} & \multirow{2}{*}{DNN} & \multirow{2}{*}{KNN} & \multicolumn{4}{c}{SOM} \\
\cline{4-5}
\cline{8-11}
&  &  & RBF & Matern &  &  & RAND &  AVG &  LS &  WAVG \\
    \hline
    B/A & 0.09 & 0.02 & 0.13 & 0.03 & 0.05 & \bf{0.01} & 2.03e2 & 1.56e2 & 0.26 & 0.11 \\
    Ash & 0.64 & 0.61 & 1.35 & 0.22 & 0.96 & \bf{0.19} & 2.43e3 & 2.01e3 & 1.32 & 0.57 \\
    BTU & 1.16e5 & 4.40e4 & 3.42e4 & 2.86e4 & 9.86e4 & 1.22e4 & 9.52e5 & 8.60e5 & 3.79e2 & \bf{1.45e2} \\
    H2O & 7.02 & 3.59 & 9.38 & 2.41 & 4.06 & \bf{1.07} & 1.04e4 & 9.83e3 & 3.38 & 1.31 \\
    Na & 2.60e6 & 2.75e6 & 4.11e6 & 1.25e6 & 2.01e6 & 2.32e5 & 7.59e6 & 6.84e6 & 1.53e3 & \bf{5.19e2} \\
    Fe & 3.10e6 & 1.94e6 & 2.12e6 & 1.99e6 & 4.70e6 & 1.07e6 & 1.27e7 & 1.14e7 & 1.84e3 & \bf{8.27e2} \\
    Al & 3.80e6 & 3.33e6 & 9.10e6 & 1.05e6 & 2.06e6 & 6.08e5 & 1.82e7 & 1.63e7 & 2.68e3 & \bf{9.75e2} \\
    Ca & 8.50e5 & 7.22e5 & 7.54e5 & 3.30e5 & 1.04e6 & 2.04e5 & 8.79e6 & 8.00e6 & 1.23e3 & \bf{5.56e2} \\
    K & 1.58e5 & 1.10e5 & 2.88e5 & 5.22e4 & 1.05e5 & 4.72e4 & 3.80e6 & 3.50e6 & 3.46e2 & \bf{1.51e2} \\
    Mg & 1.37e5 & 4.26e4 & 2.29e5 & 5.08e4 & 1.45e5 & 2022e4 & 5.39e6 & 4.95e6 & 4.60e2 & \bf{1.88e2} \\
    Si & 2.66e7 & 2.90e7 & 6.55e7 & 6.35e6 & 4.07e7 & 4.10e6 & 1.08e8 & 9.88e7 & 8.82e3 & \bf{3.48e3} \\
    SO & 1.46e7 & 4.51e6 & 5.04e6 & 4.56e6 & 1.45e7 & 2.20e6 &4.32e7 & 3.98e7 & 3.46e3 & \bf{1.48e3} \\
    Ti & 1.26e4 & 8.43e3 & 2.48e4 & 3.05e3 & 6.05e3 & 9.63e2 & 1.43e6 & 1.33e6 & 1.17e2 & \bf{45.16} \\
    \hline
    Average & 4.00e6 & 3.27e6 & 6.71e6 & 1.20e6 & 5.03e6 & 6.54e5 &1.62e7 & 1.48e7 & 1.61e3 & \bf{6.44e2} \\

\hline
\end{tabular}
\end{table*}

\begin{table*}[!t]
  \centering
  \caption{Energy Data RMSE}
  \scriptsize
  \label{tab:energy}
  \vspace{-0.2cm}
  \begin{tabular}{c|*{3}{c}|*{3}{c}|*{3}{c}}
  \hline
  & \multicolumn{3}{c|}{50} & \multicolumn{3}{c|}{100} & \multicolumn{3}{c}{200} \\
    \hline
    & Minmax & Standard & Robust & Minmax & Standard & Robust & Minmax & Standard & Robust\\
    \hline
    Linear & 1.72e4 & 1.72e4 & 1.72e4 & 9.05e3 & 9.05e3 & 9.05e3 & 7.50e3 & 7.50e3 & 7.50e3 \\
    Poly &  7.85e3 & 5.49e3 & 1.01e4 & 1.23e4 & 1.20e4 & 1.59e4 & 1.03e4 & 9.93e3 & 1.10e4\\
    \hline
    DNN &  5.35e3 & 5.58e3 & 5.25e3 & 7.35e3 & 7.73e3 & 7.03e3 & 6.52e3 & 6.36e3 & 6.44e3 \\
    \hline
    KNN & 5.44e3 & 6.65e3 & 6.26e3 & 6.90e3 & 6.84e3 & 7.01e3 & 5.71e3 & 6.02e3 & 6.41e3 \\
    \hline
    GPR(RBF) & 5.93e3 & 5.10e3 & 6.23e3 & 7.25e3 & 6.97e3 & 7.82e3 & 5.81e3 & 5.48e3 & 6.25e3\\
    GPR(Matern) & 5.06e3 & 4.83e3 & 6.09e3 & 7.11e3 & 6.82e3 & 7.20e3 & 5.78e3 & 5.91e3 & 5.75e3\\
    \hline
    SOM WAVG & \bf{8.32e1} & \bf{8.57e1} & \bf{8.04e1} & \bf{8.75e1} & \bf{8.69e1} & \bf{8.48e1} & \bf{9.01e1} & \bf{9.25e1} & \bf{9.14e1}\\
    
    \hline
  \end{tabular}
  \vspace{-0.25cm}
\end{table*}

  
    

\begin{table}[!t]
  \centering
  \caption{GPR and DNN testing RMSE using $1600$ training samples.}
  \vspace{-0.2cm}
  \scriptsize
  \label{tab:energy_1600}
  \begin{tabular}{c|*{3}{c}}
    \hline
    & Minmax & Standard & Robust \\
    \hline
    GPR(RBF) & 6.81e3 & 6.85e3 & 7.82e3 \\
    GPR(Matern) & 6.68e3 & 6.74e3 & 7.54e3 \\
    \hline
    DNN & 8.38e3 & 1.20e4 & 8.24e3 \\ 
    \hline
  \end{tabular}
  \vspace{-0.25cm}
\end{table}

\subsection{Topological Regression vs Regression Methods}
Our proposed method uses very few labeled data points for unknown data estimation. Generally, other supervised methods perform very poor when training data examples are limited and input data is high dimensional. In these experiments, to get a sense of how these baseline methods perform, we tested supervised regressions method such as linear regression, polynomial regression, GPR, DNN, and KNN. 


Tables~\ref{tab:random_guess}, and~\ref{tab:minmax_rmse} show the validation prediction RMSE in the original scale of each coal (data) property, across all the experimental methodologies. Experiments are conducted using all three data normalization methods: Minmax, Standard, and Robust. Table~\ref{tab:minmax_rmse} shows the results using Minmax normalization, using Standard and Robust normalization shows similar results as Minmax. The results of using Standard and Robust normalization can be found in the Appendix. Table~\ref{tab:random_guess} provides a baseline, with the first column $X \sim U$ showing RSME using random predictions based on the uniform distribution of each coal property's value range, and the second column $X \sim N$ shows the random predictions using the normal distribution of each coal property data value. As the original dataset contains an extremely limited number of labeled data points for validation of the regression methods, the validation RMSE of those methods can be very high. We use random guessing RMSE to show if a method is valid at making predictions, i.e., do they beat random guessing. We then further compare the SOM topological projections to powerful methods including Gaussian processes (GPs) and DNNs.

\paragraph{Classical Methods}
We tested two classical methods, linear regression and polynomial regression. Notice that the prediction performance of these methods is close to random guessing. Polynomial regression performs slightly better than linear regression with min-max normalization while linear regression performs better using the other two normalization methods. Since the results show the RMSE of each coal property value's in the unnormalized data scale, depicting the average performance across all coal properties is biased to parameters with higher value range, we do not over-speculate on the average RMSE provided in the bottom row.
\paragraph{Gaussian Process Regression (GPR)}
is a non-parametric, Bayesian approach to regression that employs a probabilistic framework for learning and inference~\cite{williams1995gaussian}, which has shown strong performance on small sized datasets.  We used two kernel functions for GPR: $Constant * RBF$ (a radial basis function kernel) and 
$Constant * Matern$ (a Matern generalized RBF kernel). 
The hyperparameter tuning range for the two kernel functions can be found in the supplementary materials. GPR regression results are generally better than random guessing and classical methods but not significantly so.

\paragraph{Deep Neural Network (DNNs)}
offer a powerful, nonlinear approach to tackling high-dimensional regression problems. After applying PCA to the input data features, we obtain $31$ (transformed) features, upon which a simple DNN is further applied to conduct regression/estimation. The structure of the DNN we crafted can be found in supplementary materials. The DNN performs better with the min-max scaling method, but generally worse than the GPR methods. As DNNs generally require large training datasets to obtain good-quality performance, this was expected.

\paragraph{K-Nearest Neighbors (KNNs)}
is a simple but widely used method for regression tasks. The Naive KNN predicts values by taking the average values of its "k" nearest neighbors. The results show KNN outperforms other methods in predicting 2-3 coal properties for coal dataset, however it generally still performs worse than our proposed methods.

\paragraph{The Self-Organizing Map.} 
To show that the distance on the U-Matrix plays an important role in parameter value estimation, we also tested the use of non-weighted average values of randomly selected neighbors, non-weighted average values of closest neighbors, as well as linear regression on closest neighbors (polynomial regression is not shown due to the poor performance we found it exhibited). The results in Table~\ref{tab:minmax_rmse} demonstrates that using the average value of random neighbors (\emph{SOM RAND}) performs the worst (as expected), using average value of closest neighbors (\emph{SOM AVG}) does better than random neighbors, and linear regression (\emph{SOM LS}
) among closest neighbors is better than average closest neighbors. Note that using the weighted average of closest neighbors (\emph{SOM WAVG}) does the best over all proposed methods as well as other regression methods. In general, we find that \emph{SOM WAVG}  performs orders of magnitude better than the other methods for almost all coal properties.


\subsection{Unsupervised DBSCAN}
It is possible to apply the idea of using the clustering result and the distance values among adjunct clusters for value estimation in other schemes beyond the SOM. In light of this, we explored using another unsupervised clustering method for the weighted average value estimation, i.e., DBSCAN (Density-Based Spatial Clustering of Applications with Noise)~\cite{ester1996density} which is an unsupervised clustering algorithm known for its ability to effectively group data points in high-dimensional spaces. DBSCAN identifies dense clusters, distinguishing them from sparser noise regions, making it highly suitable for high-dimensional datasets.

\begin{figure}[!t]
  \centering
  \includegraphics[width=\columnwidth]{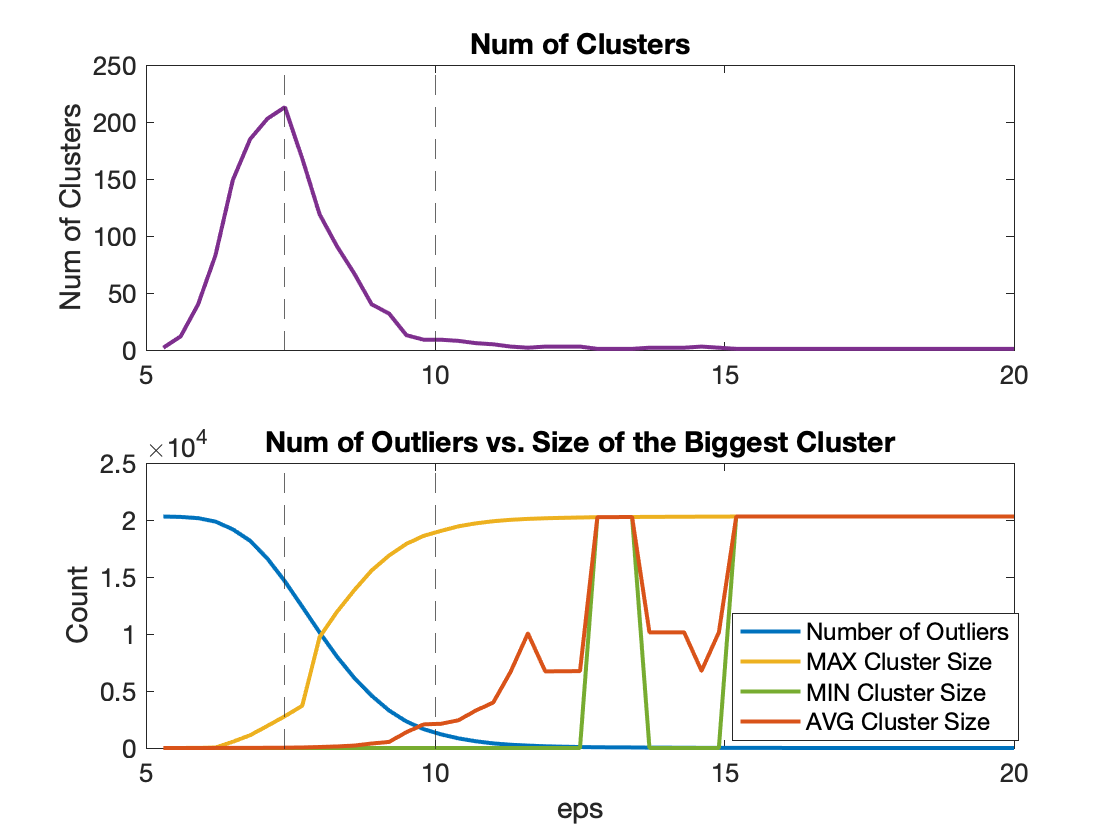}
  \vspace{-0.2cm}
  \caption{DBSCAN with PCA.}
  \label{fig:dbscan_pca}
  \vspace{-0.25cm}
\end{figure}

DBSCAN requires two main parameters: $eps$ and $min\_samples$. $eps$ is the maximum distance between two samples for one to be considered as in the neighborhood of the other. $min\_samples$ is the number of samples in a neighborhood for a point to be considered as a core point; this includes the point itself. The plot below show the number of clusters and cluster size using $min\_samples = 3$ and different $eps$. Figure \ref{fig:dbscan_pca} show the relationship of using different $eps$ and the number of clusters, size of the biggest, average, and smallest cluster, and the number of outliers on the original data and data with $80$\% variance preserved utilizing PCA.

The vertical dash line shows the potential $eps$ for clustering and its corresponding number of clusters and the number of outliers. When it reaches the highest number of total clusters, the amount of outliers are around $20$K and all clusters are extremely small. When the number of outliers drops significantly, the biggest cluster contains almost all the datapoints, which left very little room for (effective) value estimation. Clustering using DBSCAN without PCA shows similar trends, and the result can be found in the Appendix. The results with DBSCAN shows that not all unsupervised learning clustering methods work with the proposed semi-supervised scheme in comparison to the SOM.


\subsection{Energy Dataset}
The coal dataset has a very limited number of labeled data points, we utilize the energy dataset to study the effectiveness of SOM topological projections using different quantities of labeled data points for training. We utilized $50$, $100$, and $200$ randomly selected labeled data points, with regression models using $80\%$ for training, and $20\%$ for validation. For the SOM topological projection, we utilized the remaining data points as unlabeled data, removing the prediction parameter target. Table \ref{tab:energy} presents the weighted average value projection compared to the other regression methods. From these results, we observe that the SOM does significantly better than the other regression methods. 

To further explore the regression capability of GPR and DNN on even larger training dataset, GPR and DNN models were trained on datasets containing $1600$ data points and tested on sets with $200$ data points. The validation results are shown in Table~\ref{tab:energy_1600}. Comparing results in Table~\ref{tab:energy} and \ref{tab:energy_1600}, even training with significantly more training data does not necessary improve the performance of complex regression models in our problem contexts. Note that the SOM topological projections, even when using a greatly limited set labeled data patterns (even just $50$ labeled samples), still outperforms these methods by several orders of magnitude.


    

    

\section{Conclusion}

This paper proposed a minimally supervised learning algorithm based on topological projections in a self-organizing map (SOM). Our scheme first clusters the unlabeled data within an SOM, then utilizes the distance between SOM units to create a graph that can be used to compute the shortest paths in the SOM topology space between nodes using Dijkstra's algorithm. Labeled data points are then mapped to the SOM where ultimately predictions of unlabeled data points can be made by utilizing topological projections based on the closest labeled nearest neighbor units from the best matching unit of the unlabeled data.

We investigated utilizing both a weighted average and regression methods to perform the parameter prediction given the nearest neighbors and found that the weighted average method has the best performance. Our results also show that our proposed SOM method performs orders of magnitude better than other supervised learning regression methods, such as classical linear and polynomial regression, Gaussian process regression, and deep neural networks, as well as an unsupervised learning scheme, such as DBSCAN. We also show that our methods still significantly outperform supervised methods even when the supervised models are provided much more labeled training data.

Our neural map method provides a promising approach for semi-supervised learning, particularly offering a viable, effective scheme for large, mostly unlabeled, datasets, particularly in instances when very few data points are labeled/annotated.  The scheme is effective with high dimensional data and further results can be easily visualized through the use of U-Matrices. Our work also provides interesting avenues for future research endeavors, such as efforts that study others methods for performing the topological projections inherent to our computational framework. Additionally, as our neural method is easily visualized via a U-Matrix, this system could be used to facilitate explainablity in model predictions.

\bibliographystyle{named}
\bibliography{99-reference}

\begin{thebibliography}{}

\bibitem[\protect\citeauthoryear{Adams \bgroup \em et al.\egroup }{2020}]{adams2020prediction}
Derrick Adams, Dong-Hoon Oh, Dong-Won Kim, Chang-Ha Lee, and Min Oh.
\newblock Prediction of sox--nox emission from a coal-fired cfb power plant with machine learning: Plant data learned by deep neural network and least square support vector machine.
\newblock {\em Journal of Cleaner Production}, 270:122310, 2020.

\bibitem[\protect\citeauthoryear{Azzalini \bgroup \em et al.\egroup }{2021}]{azzalini2021minimally}
Davide Azzalini, Luca Bonali, and Francesco Amigoni.
\newblock A minimally supervised approach based on variational autoencoders for anomaly detection in autonomous robots.
\newblock {\em IEEE Robotics and Automation Letters}, 6(2):2985--2992, 2021.

\bibitem[\protect\citeauthoryear{Barnes \bgroup \em et al.\egroup }{2016}]{barnes2016real}
Sean Barnes, Eric Hamrock, Matthew Toerper, Sauleh Siddiqui, and Scott Levin.
\newblock Real-time prediction of inpatient length of stay for discharge prioritization.
\newblock {\em Journal of the American Medical Informatics Association}, 23(e1):e2--e10, 2016.

\bibitem[\protect\citeauthoryear{Candanedo \bgroup \em et al.\egroup }{2017}]{candanedo2017data}
Luis~M Candanedo, V{\'e}ronique Feldheim, and Dominique Deramaix.
\newblock Data driven prediction models of energy use of appliances in a low-energy house.
\newblock {\em Energy and buildings}, 140:81--97, 2017.

\bibitem[\protect\citeauthoryear{Che \bgroup \em et al.\egroup }{2012}]{che2012adaptive}
Jinxing Che, Jianzhou Wang, and Guangfu Wang.
\newblock An adaptive fuzzy combination model based on self-organizing map and support vector regression for electric load forecasting.
\newblock {\em Energy}, 37(1):657--664, 2012.

\bibitem[\protect\citeauthoryear{Dai \bgroup \em et al.\egroup }{2023}]{dai2023semi}
Weihang Dai, Xiaomeng Li, and Kwang-Ting Cheng.
\newblock Semi-supervised deep regression with uncertainty consistency and variational model ensembling via bayesian neural networks.
\newblock {\em arXiv preprint arXiv:2302.07579}, 2023.

\bibitem[\protect\citeauthoryear{de Souza~Junior \bgroup \em et al.\egroup }{2015}]{de2015minimal}
Amauri~Holanda de~Souza~Junior, Francesco Corona, Guilherme~A Barreto, Yoan Miche, and Amaury Lendasse.
\newblock Minimal learning machine: A novel supervised distance-based approach for regression and classification.
\newblock {\em Neurocomputing}, 164:34--44, 2015.

\bibitem[\protect\citeauthoryear{Dong \bgroup \em et al.\egroup }{2015}]{dong2015novel}
Zibo Dong, Dazhi Yang, Thomas Reindl, and Wilfred~M Walsh.
\newblock A novel hybrid approach based on self-organizing maps, support vector regression and particle swarm optimization to forecast solar irradiance.
\newblock {\em Energy}, 82:570--577, 2015.

\bibitem[\protect\citeauthoryear{Ester \bgroup \em et al.\egroup }{1996}]{ester1996density}
Martin Ester, Hans-Peter Kriegel, J{\"o}rg Sander, Xiaowei Xu, et~al.
\newblock A density-based algorithm for discovering clusters in large spatial databases with noise.
\newblock In {\em kdd}, volume~96, pages 226--231, 1996.

\bibitem[\protect\citeauthoryear{Hsu \bgroup \em et al.\egroup }{2009}]{hsu2009two}
Sheng-Hsun Hsu, JJ~Po-An Hsieh, Ting-Chih Chih, and Kuei-Chu Hsu.
\newblock A two-stage architecture for stock price forecasting by integrating self-organizing map and support vector regression.
\newblock {\em Expert Systems with Applications}, 36(4):7947--7951, 2009.

\bibitem[\protect\citeauthoryear{Jiang \bgroup \em et al.\egroup }{2021a}]{jiang2021semi}
Jehn-Ruey Jiang, Jian-Bin Kao, and Yu-Lin Li.
\newblock Semi-supervised time series anomaly detection based on statistics and deep learning.
\newblock {\em Applied Sciences}, 11(15):6698, 2021.

\bibitem[\protect\citeauthoryear{Jiang \bgroup \em et al.\egroup }{2021b}]{jiang2021deep}
Yiqi Jiang, Chaolin Li, Lu~Sun, Dong Guo, Yituo Zhang, and Wenhui Wang.
\newblock A deep learning algorithm for multi-source data fusion to predict water quality of urban sewer networks.
\newblock {\em Journal of Cleaner Production}, 318:128533, 2021.

\bibitem[\protect\citeauthoryear{Kohonen \bgroup \em et al.\egroup }{1996}]{kohonen1996engineering}
Teuvo Kohonen, Erkki Oja, Olli Simula, Aari Visa, and Jari Kangas.
\newblock Engineering applications of the self-organizing map.
\newblock {\em Proceedings of the IEEE}, 84(10):1358--1384, 1996.

\bibitem[\protect\citeauthoryear{Kohonen}{1982}]{kohonen1982self}
Teuvo Kohonen.
\newblock Self-organized formation of topologically correct feature maps.
\newblock {\em Biological cybernetics}, 43(1):59--69, 1982.

\bibitem[\protect\citeauthoryear{Kohonen}{1990}]{kohonen1990self}
Teuvo Kohonen.
\newblock The self-organizing map.
\newblock {\em Proceedings of the IEEE}, 78(9):1464--1480, 1990.

\bibitem[\protect\citeauthoryear{Kohonen}{2013}]{kohonen2013essentials}
Teuvo Kohonen.
\newblock Essentials of the self-organizing map.
\newblock {\em Neural networks}, 37:52--65, 2013.

\bibitem[\protect\citeauthoryear{Kostopoulos \bgroup \em et al.\egroup }{2019}]{kostopoulos2019semi}
Georgios Kostopoulos, Sotiris Kotsiantis, Nikos Fazakis, Giannis Koutsonikos, and Christos Pierrakeas.
\newblock A semi-supervised regression algorithm for grade prediction of students in distance learning courses.
\newblock {\em International Journal on Artificial Intelligence Tools}, 28(04):1940001, 2019.

\bibitem[\protect\citeauthoryear{Kure \bgroup \em et al.\egroup }{2022}]{kure2022asset}
Halima~Ibrahim Kure, Shareeful Islam, Mustansar Ghazanfar, Asad Raza, and Maruf Pasha.
\newblock Asset criticality and risk prediction for an effective cybersecurity risk management of cyber-physical system.
\newblock {\em Neural Computing and Applications}, 34(1):493--514, 2022.

\bibitem[\protect\citeauthoryear{Levati{\'c} \bgroup \em et al.\egroup }{2015}]{levatic2015semi}
Jurica Levati{\'c}, Michelangelo Ceci, Dragi Kocev, and Sa{\v{s}}o D{\v{z}}eroski.
\newblock Semi-supervised learning for multi-target regression.
\newblock In {\em New Frontiers in Mining Complex Patterns: Third International Workshop, NFMCP 2014, Held in Conjunction with ECML-PKDD 2014, Nancy, France, September 19, 2014, Revised Selected Papers 3}, pages 3--18. Springer, 2015.

\bibitem[\protect\citeauthoryear{Li \bgroup \em et al.\egroup }{2017}]{li2017learning}
Yu-Feng Li, Han-Wen Zha, and Zhi-Hua Zhou.
\newblock Learning safe prediction for semi-supervised regression.
\newblock In {\em Proceedings of the AAAI Conference on Artificial Intelligence}, volume~31, 2017.

\bibitem[\protect\citeauthoryear{Meattini \bgroup \em et al.\egroup }{2022}]{meattini2022semg}
Roberto Meattini, Alessandra Bernardini, Gianluca Palli, and Claudio Melchiorri.
\newblock semg-based minimally supervised regression using soft-dtw neural networks for robot hand grasping control.
\newblock {\em IEEE Robotics and Automation Letters}, 7(4):10144--10151, 2022.

\bibitem[\protect\citeauthoryear{Qiang \bgroup \em et al.\egroup }{2023}]{qiang2023minimally}
Chunyu Qiang, Hao Li, Hao Ni, He~Qu, Ruibo Fu, Tao Wang, Longbiao Wang, and Jianwu Dang.
\newblock Minimally-supervised speech synthesis with conditional diffusion model and language model: A comparative study of semantic coding.
\newblock {\em arXiv preprint arXiv:2307.15484}, 2023.

\bibitem[\protect\citeauthoryear{Rezagholiradeh and Haidar}{2018}]{rezagholiradeh2018reg}
Mehdi Rezagholiradeh and Md~Akmal Haidar.
\newblock Reg-gan: Semi-supervised learning based on generative adversarial networks for regression.
\newblock In {\em 2018 IEEE international conference on acoustics, speech and signal processing (ICASSP)}, pages 2806--2810. IEEE, 2018.

\bibitem[\protect\citeauthoryear{Riese \bgroup \em et al.\egroup }{2019}]{riese2019supervised}
Felix~M Riese, Sina Keller, and Stefan Hinz.
\newblock Supervised and semi-supervised self-organizing maps for regression and classification focusing on hyperspectral data.
\newblock {\em Remote Sensing}, 12(1):7, 2019.

\bibitem[\protect\citeauthoryear{Silva and Del-Moral-Hernandez}{2011}]{silva2011som}
Leandro~A Silva and Emilio Del-Moral-Hernandez.
\newblock A som combined with knn for classification task.
\newblock In {\em The 2011 International Joint Conference on Neural Networks}, pages 2368--2373. IEEE, 2011.

\bibitem[\protect\citeauthoryear{Singh and Singh}{2020}]{singh2020investigating}
Dalwinder Singh and Birmohan Singh.
\newblock Investigating the impact of data normalization on classification performance.
\newblock {\em Applied Soft Computing}, 97:105524, 2020.

\bibitem[\protect\citeauthoryear{Smrekar \bgroup \em et al.\egroup }{2010}]{smrekar2010prediction}
J~Smrekar, D~Pandit, Magnus Fast, Mohsen Assadi, and Sudipta De.
\newblock Prediction of power output of a coal-fired power plant by artificial neural network.
\newblock {\em Neural Computing and Applications}, 19:725--740, 2010.

\bibitem[\protect\citeauthoryear{Suchenwirth \bgroup \em et al.\egroup }{2014}]{suchenwirth2014large}
Leonhard Suchenwirth, Wolfgang St{\"u}mer, Tobias Schmidt, Michael F{\"o}rster, and Birgit Kleinschmit.
\newblock Large-scale mapping of carbon stocks in riparian forests with self-organizing maps and the k-nearest-neighbor algorithm.
\newblock {\em Forests}, 5(7):1635--1652, 2014.

\bibitem[\protect\citeauthoryear{Thakkar and Chaudhari}{2021}]{thakkar2021fusion}
Ankit Thakkar and Kinjal Chaudhari.
\newblock Fusion in stock market prediction: a decade survey on the necessity, recent developments, and potential future directions.
\newblock {\em Information Fusion}, 65:95--107, 2021.

\bibitem[\protect\citeauthoryear{Tian \bgroup \em et al.\egroup }{2014}]{tian2014anomaly}
Jing Tian, Michael~H Azarian, and Michael Pecht.
\newblock Anomaly detection using self-organizing maps-based k-nearest neighbor algorithm.
\newblock In {\em PHM society European conference}, volume~2, 2014.

\bibitem[\protect\citeauthoryear{Vesanto and Alhoniemi}{2000}]{vesanto2000clustering}
Juha Vesanto and Esa Alhoniemi.
\newblock Clustering of the self-organizing map.
\newblock {\em IEEE Transactions on neural networks}, 11(3):586--600, 2000.

\bibitem[\protect\citeauthoryear{Villa-P{\'e}rez \bgroup \em et al.\egroup }{2021}]{villa2021semi}
Miryam~Elizabeth Villa-P{\'e}rez, Miguel~A Alvarez-Carmona, Octavio Loyola-Gonzalez, Miguel~Angel Medina-P{\'e}rez, Juan~Carlos Velazco-Rossell, and Kim-Kwang~Raymond Choo.
\newblock Semi-supervised anomaly detection algorithms: A comparative summary and future research directions.
\newblock {\em Knowledge-Based Systems}, 218:106878, 2021.

\bibitem[\protect\citeauthoryear{Wang \bgroup \em et al.\egroup }{2020a}]{wang2020focalmix}
Dong Wang, Yuan Zhang, Kexin Zhang, and Liwei Wang.
\newblock Focalmix: Semi-supervised learning for 3d medical image detection.
\newblock In {\em Proceedings of the IEEE/CVF Conference on Computer Vision and Pattern Recognition}, pages 3951--3960, 2020.

\bibitem[\protect\citeauthoryear{Wang \bgroup \em et al.\egroup }{2020b}]{wang2020semi}
Yaxing Wang, Salman Khan, Abel Gonzalez-Garcia, Joost van~de Weijer, and Fahad~Shahbaz Khan.
\newblock Semi-supervised learning for few-shot image-to-image translation.
\newblock In {\em Proceedings of the IEEE/CVF Conference on Computer Vision and Pattern Recognition}, pages 4453--4462, 2020.

\bibitem[\protect\citeauthoryear{Wang \bgroup \em et al.\egroup }{2020c}]{wang2020elucidating}
Zhan Wang, Jun Xiao, Lingqing Wang, Tao Liang, Qingjun Guo, Yunlan Guan, and Joerg Rinklebe.
\newblock Elucidating the differentiation of soil heavy metals under different land uses with geographically weighted regression and self-organizing map.
\newblock {\em Environmental Pollution}, 260:114065, 2020.

\bibitem[\protect\citeauthoryear{Williams and Rasmussen}{1995}]{williams1995gaussian}
Christopher Williams and Carl Rasmussen.
\newblock Gaussian processes for regression.
\newblock {\em Advances in neural information processing systems}, 8, 1995.

\bibitem[\protect\citeauthoryear{Zhang \bgroup \em et al.\egroup }{2021}]{zhang2021minimally}
Xinyang Zhang, Chenwei Zhang, Xin~Luna Dong, Jingbo Shang, and Jiawei Han.
\newblock Minimally-supervised structure-rich text categorization via learning on text-rich networks.
\newblock In {\em Proceedings of the Web Conference 2021}, pages 3258--3268, 2021.

\end{thebibliography}

\appendix

\newpage

\section{Experimental Details}

\subsection{SOM Hyperparameters}
Table~\ref{tab:som_hp}~shows the hyperparameters used for all SOM (self-organizing map) experiments.

\begin{table}[h]
\centering
\footnotesize
\caption{SOM hyperparameters}
\label{tab:som_hp}
\begin{tabular}{l|c}
\hline
Learning Rate Start & 0.5 \\
\hline
Learning Rate End & 0.05 \\
\hline 
Radius MAX & $max(N\textrm{-}Columns, N\textrm{-}Rows)/2$ \\
\hline
Radius MIN & 1 \\
\hline
Neighborhood Function & Gaussian \\
\hline
Train Epochs & 20,000 \\
\hline
Initialization Method & Random \\
\hline
Distance Metric & Euclidean \\

\hline
\end{tabular}
\end{table}

\subsection{Gaussion Process Regression}

Table~\ref{tab:gpr}~shows the GPR (Gaussion process regression)  model initial hyperparameter range for the coal and energy dataset experiments.

\begin{table}[h]
\centering
\caption{GPR Initial Hyperparameter Range}
\label{tab:gpr}
\scriptsize
\begin{tabular}{c|c|c|c|c}
\hline
Kernal & $\alpha$ & C & Length Scale & Nu \\
\hline
$C \cdot RBF$ & [1e-5, 1e-1] & [1e-5, 1e1] & [1e-1, 10.0] & \ \\
$C \cdot Matern$ & [1e-5, 1e-1] & [1e-5, 1e1] & [1e-1, 10.0] & [0.1, 2.5] \\
\hline
\end{tabular}
\end{table}

\subsection{Deep Neural Networks}
Table~\ref{tab:dnn_coal} shows the DNN (deep neural network) structure for the coal dataset. The coal dataset has 31 features after PCA as inputs to the DNN.
Table~\ref{tab:dnn_energy}~shows the DNN structure for the energy dataset. The energy dataset has 27 input parameters to the DNN.

\begin{table}[t]
\centering
\caption{DNN Model Structure}
\label{tab:dnn_coal}
\begin{tabular}{lll}
\hline
Layer (type) & Output Shape & Activation \\
\hline
Dense (31) & (None, 31) & ReLU \\
Dense (15) & (None, 15) & ReLU \\
Dense (10) & (None, 10) & ReLU \\
Dense (1) & (None, 1) & Linear \\
\hline
\end{tabular}
\end{table}

\begin{table}[h]
\centering
\caption{DNN Model Structure}
\label{tab:dnn_energy}
\begin{tabular}{lll}
\hline
Layer (type) & Output Shape & Activation \\
\hline
Dense (27) & (None, 27) & ReLU \\
Dense (15) & (None, 15) & ReLU \\
Dense (10) & (None, 10) & ReLU \\
Dense (1) & (None, 1) & Linear \\
\hline
\end{tabular}
\end{table}

\begin{table}[t]
\centering
\footnotesize
\caption{Validation RMSE of SOM Linear Line Search}
\label{tab:som_linear}
\begin{tabular}{c|*{6}{c}}
\hline
    & N=3 & N=5 & N=7 & N=10 & N=12 & N=15 \\
\hline
\multicolumn{7}{c}{Minmax Normalization} \\
\hline
10x10 & 5.34e4  & 6.44e3  & 1.44e3  & 1.09e3  & 9.52e2  & 9.45e2 \\
15x15 & 1.51e4  & 1.92e3  & 1.52e3  & 1.35e3  & 1.27e3  & 1.18e3 \\
20x20 & 4.98e3  & 1.75e3  & 1.28e3  & 9.98e2  & 9.34e2  & 9.15e2 \\
25x25 & 3.25e4  & 1.46e3  & 1.14e3  & 1.00e3  & 8.68e2  & \bf{7.75e2} \\
30x30 & 8.16e4  & 1.61e3  & 1.44e3  & 1.57e3  & 1.34e3  & 1.23e3 \\
\hline
\multicolumn{7}{c}{Standard Normalization} \\
\hline
10x10 & 1.05e4  & 5.62e4  & 1.60e3  & 1.17e3  & 1.01e3  & 8.91e2 \\
15x15 & 1.04e4  & 2.10e3  & 1.55e3  & 1.31e3  & 1.21e3  & 1.14e3 \\
20x20 & 6.56e3  & 2.64e3  & 1.55e3  & 1.13e3  & 9.91e2  & 9.09e2 \\
25x25 & 5.79e3  & 1.71e3  & 1.21e3  & 9.82e2  & 9.63e2  & 8.81e2 \\
30x30 & 6.58e3  & 1.76e3  & 1.38e3  & 8.62e2  & 8.02e2  & \bf{7.62e2} \\
\hline
\multicolumn{7}{c}{Robust Normalization} \\
\hline
10x10 & 9.69e3  & 7.49e3  & 2.77e3  & 1.25e3  & 9.98e2  & 9.88e2 \\
15x15 & 8.72e3  & 2.05e3  & 1.45e3  & 1.25e3  & 1.16e3  & 1.09e3 \\
20x20 & 9.06e3  & 2.11e3  & 1.20e3  & 9.51e2  & 8.83e2  & 8.52e2 \\
25x25 & 5.07e3  & 1.45e3  & 1.16e3  & 1.09e3  & 8.59e2  & \bf{7.90e2} \\
30x30 & 9.71e4  & 2.13e3  & 1.38e3  & 1.73e3  & 1.86e3  & 1.60e3 \\
\hline
\end{tabular}
\end{table}

\begin{table}[t]
\centering
\footnotesize
\caption{Validation RMSE of SOM Polynomial Line Search}
\label{tab:som_poly}
\begin{tabular}{c|*{6}{c}}
\hline
    & N=3 & N=5 & N=7 & N=10 & N=12 & N=15 \\
\hline
\multicolumn{7}{c}{Minmax Normalization} \\
\hline
10x10 & 2.46e16  & 2.86e16  & 1.19e5  & 8.70e3  & 3.25e3  & 2.26e3 \\
15x15 & 2.22e17  & 1.86e15  & 1.75e4  & 4.39e3  & 3.11e3  & 3.02e3 \\
20x20 & 3.02e5  & 2.50e4  & 4.91e3  & 2.94e3  & 2.83e3  & 2.00e3 \\
25x25 & 1.20e15  & 3.42e4  & 1.06e4  & 3.37e3  & 3.41e3  & \bf{1.77e3} \\
30x30 & 1.25e16  & 2.56e4  & 1.20e4  & 7.73e3  & 4.31e3  & 3.42e3 \\    
\hline
\multicolumn{7}{c}{Standard Normalization} \\
\hline
10x10 & 8.95e15  & 3.15e16  & 3.52e15  & 5.30e3  & 3.37e3  & 2.16e3 \\
15x15 & 1.79e16  & 4.12e4  & 1.20e4  & 5.03e3  & 3.86e3  & 2.74e3 \\
20x20 & 8.55e14  & 1.43e5  & 1.97e4  & 8.17e3  & 3.95e3  & 2.63e3 \\
25x25 & 8.62e16  & 2.15e4  & 7.58e3  & 2.84e3  & 3.19e3  & 2.19e3 \\
30x30 & 1.67e15  & 2.94e4  & 9.03e3  & 2.14e3  & 1.83e3  & \bf{1.48e3} \\
\hline
\multicolumn{7}{c}{Robust Normalization} \\
\hline
10x10 & 4.44e16  & 7.72e16  & 8.59e4  & 7.86e3  & 8.01e3  & 2.66e3 \\
15x15 & 9.23e16  & 4.97e4  & 1.40e4  & 5.82e3  & 4.29e3  & 2.85e3 \\
20x20 & 4.12e16  & 8.97e4  & 9.82e3  & 3.03e3  & 2.60e3  & \bf{1.65e3} \\
25x25 & 7.35e5  & 1.25e4  & 8.99e3  & 3.59e3  & 2.23e3  & 1.78e3 \\
30x30 & 1.83e16  & 1.09e5  & 9.48e3  & 1.18e4  & 1.08e4  & 4.33e3 \\
\hline

\end{tabular}
\end{table}

\section{Additional Results}
\subsection{SOM Line Search}

Table~\ref{tab:som_linear}~shows the full results of topological projection using linear regression of the SOM closest neighbors. Table~\ref{tab:som_poly}~shows the full results of topological projection using polynomial regression of the SOM closest neighbors.

\subsection{SOM vs. GPR Prediction Plots}

\begin{figure*}[htbp]
  \centering
  \begin{subfigure}[b]{0.45\textwidth}
    \centering
    \includegraphics[width=\textwidth]{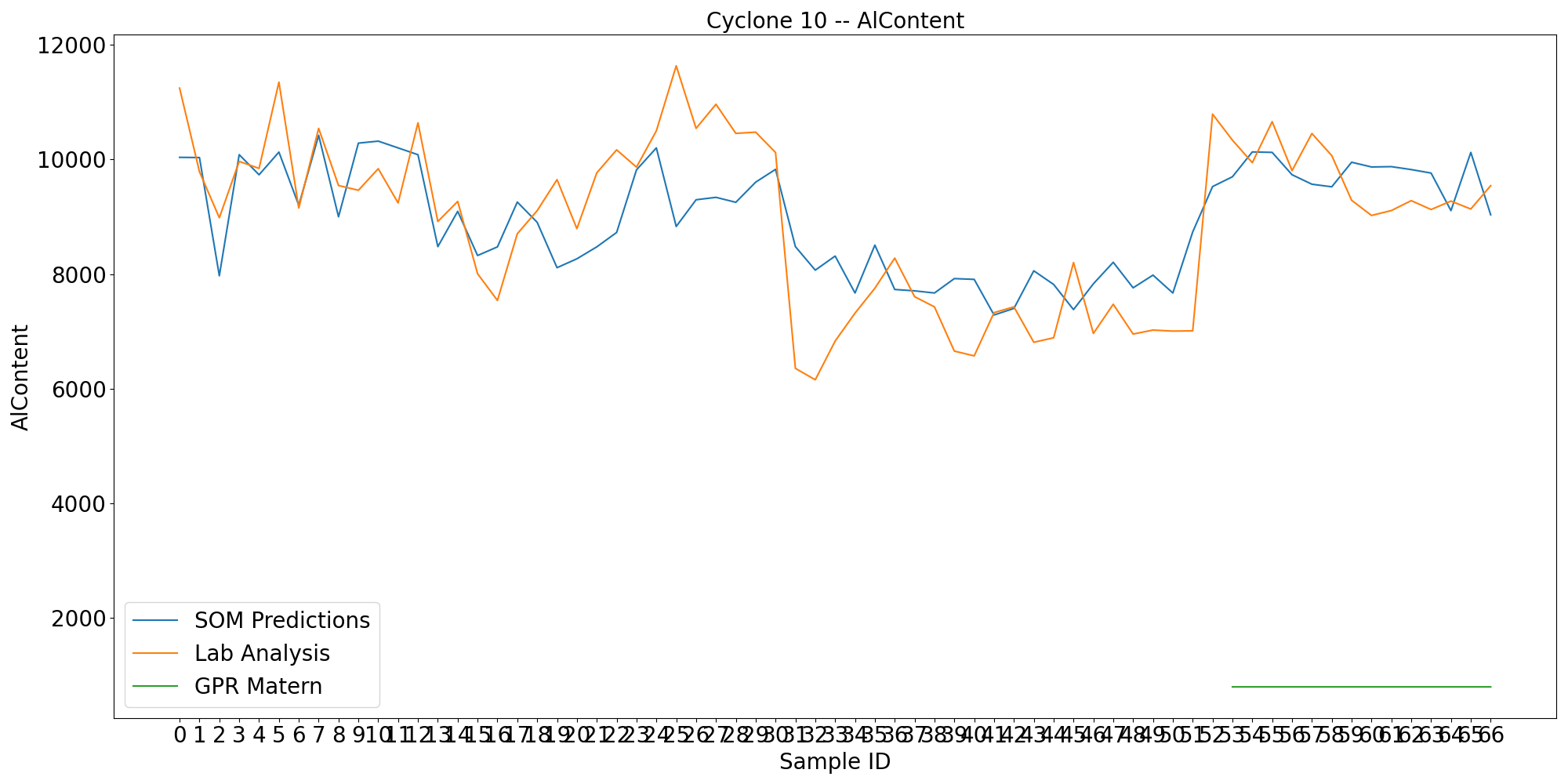}
    \caption{Al Content}
    \label{subfig:plot1}
  \end{subfigure}
  \hfill
  \begin{subfigure}[b]{0.45\textwidth}
    \centering
    \includegraphics[width=\textwidth]{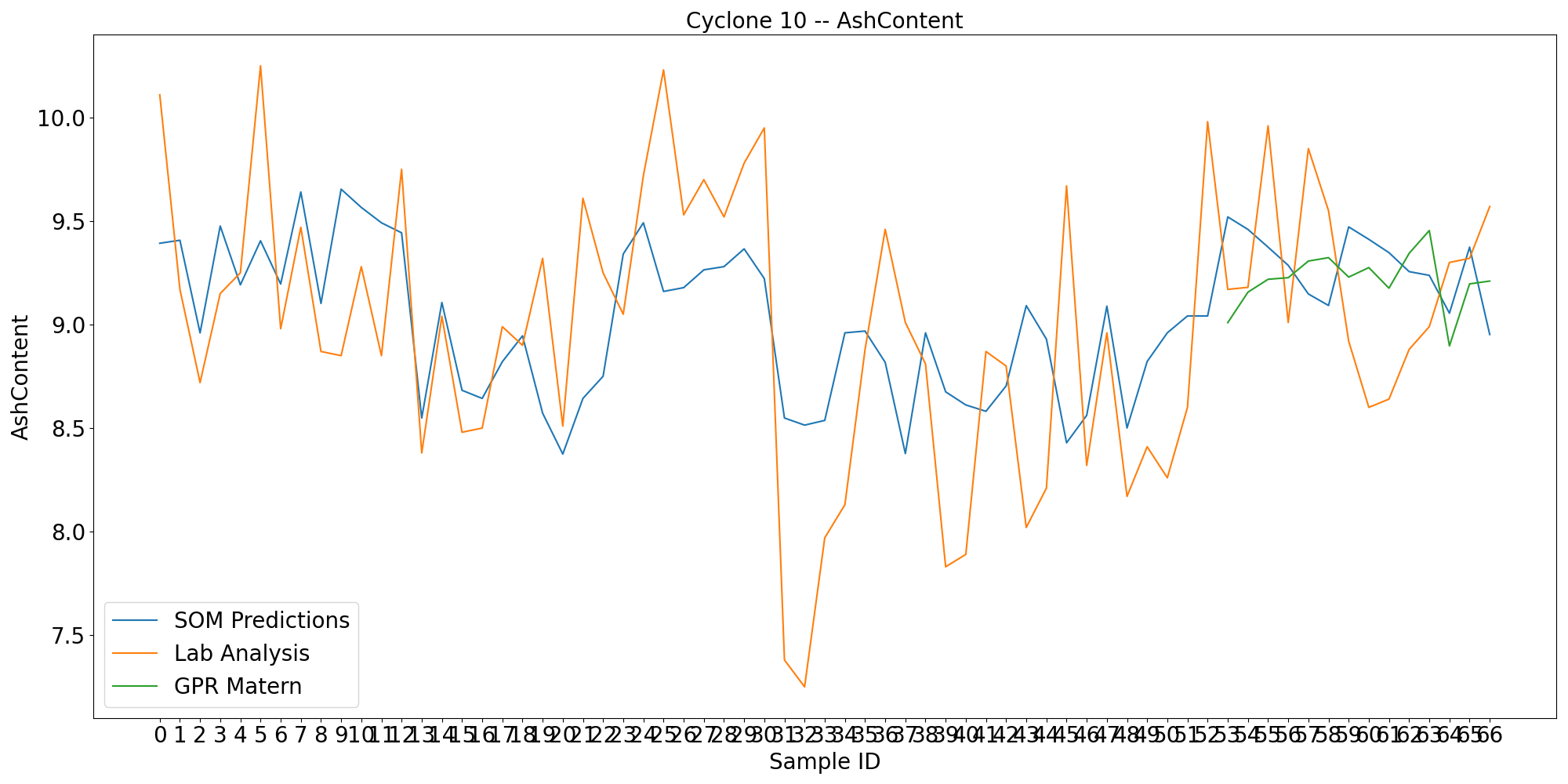}
    \caption{Ash Content}
    \label{subfig:plot2}
  \end{subfigure}
  
  \vspace{0.5cm}
  
  \begin{subfigure}[b]{0.45\textwidth}
    \centering
    \includegraphics[width=\textwidth]{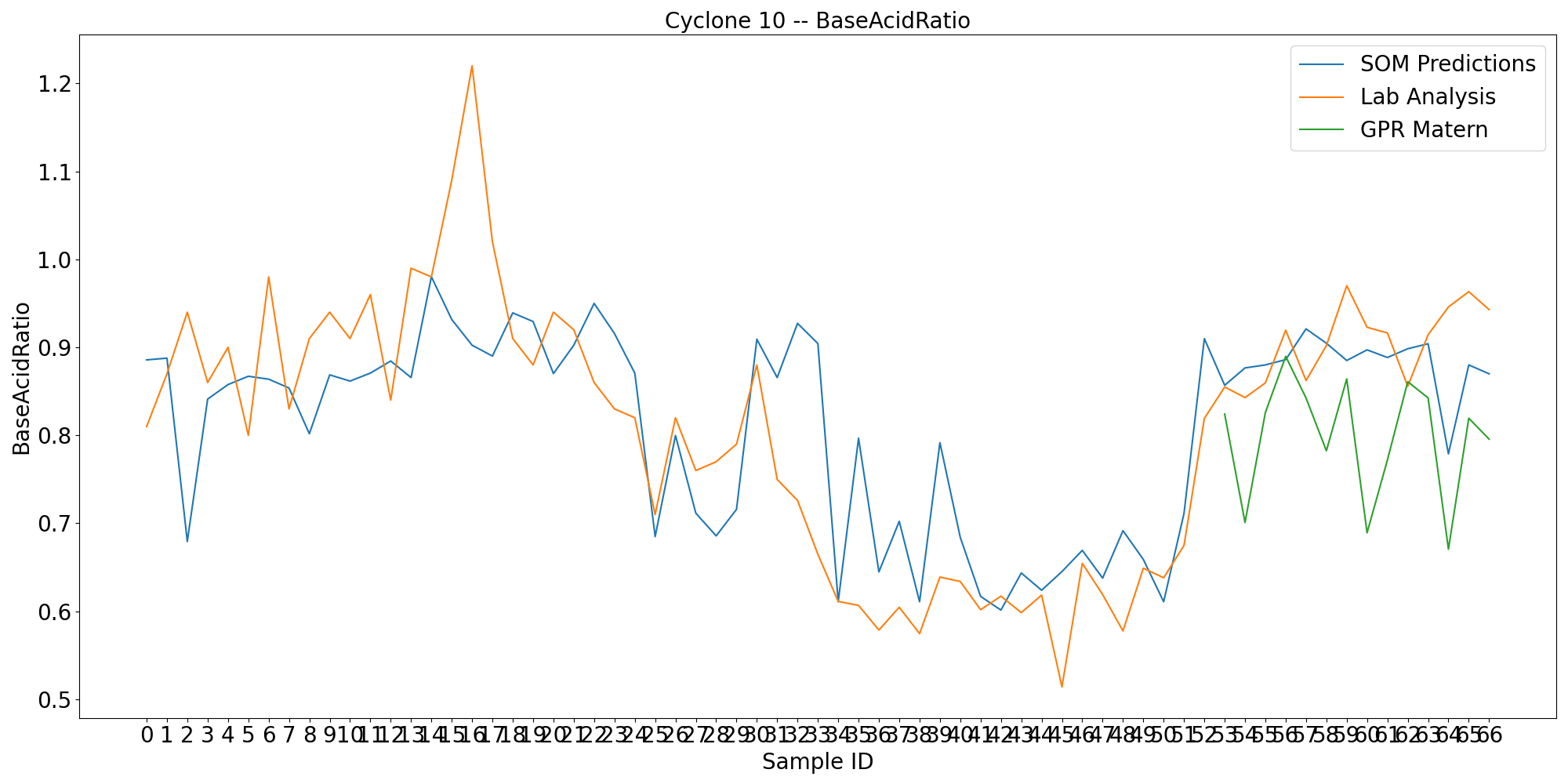}
    \caption{Base Acid Ratio}
    \label{subfig:plot3}
  \end{subfigure}
  \hfill
  \begin{subfigure}[b]{0.45\textwidth}
    \centering
    \includegraphics[width=\textwidth]{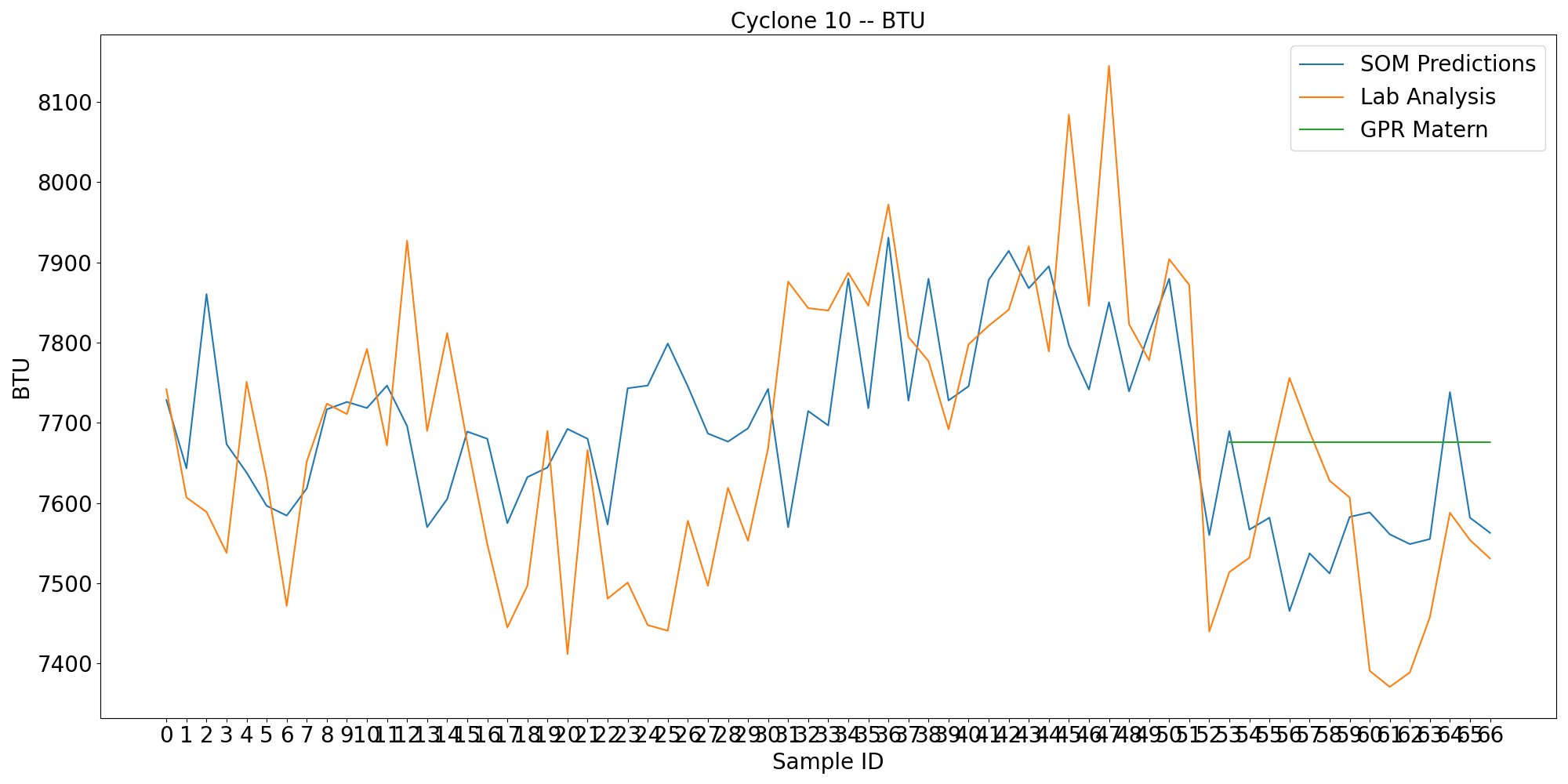}
    \caption{Na Content}
    \label{subfig:plot4}
  \end{subfigure}

  \vspace{0.5cm}
  
  \begin{subfigure}[b]{0.45\textwidth}
    \centering
    \includegraphics[width=\textwidth]{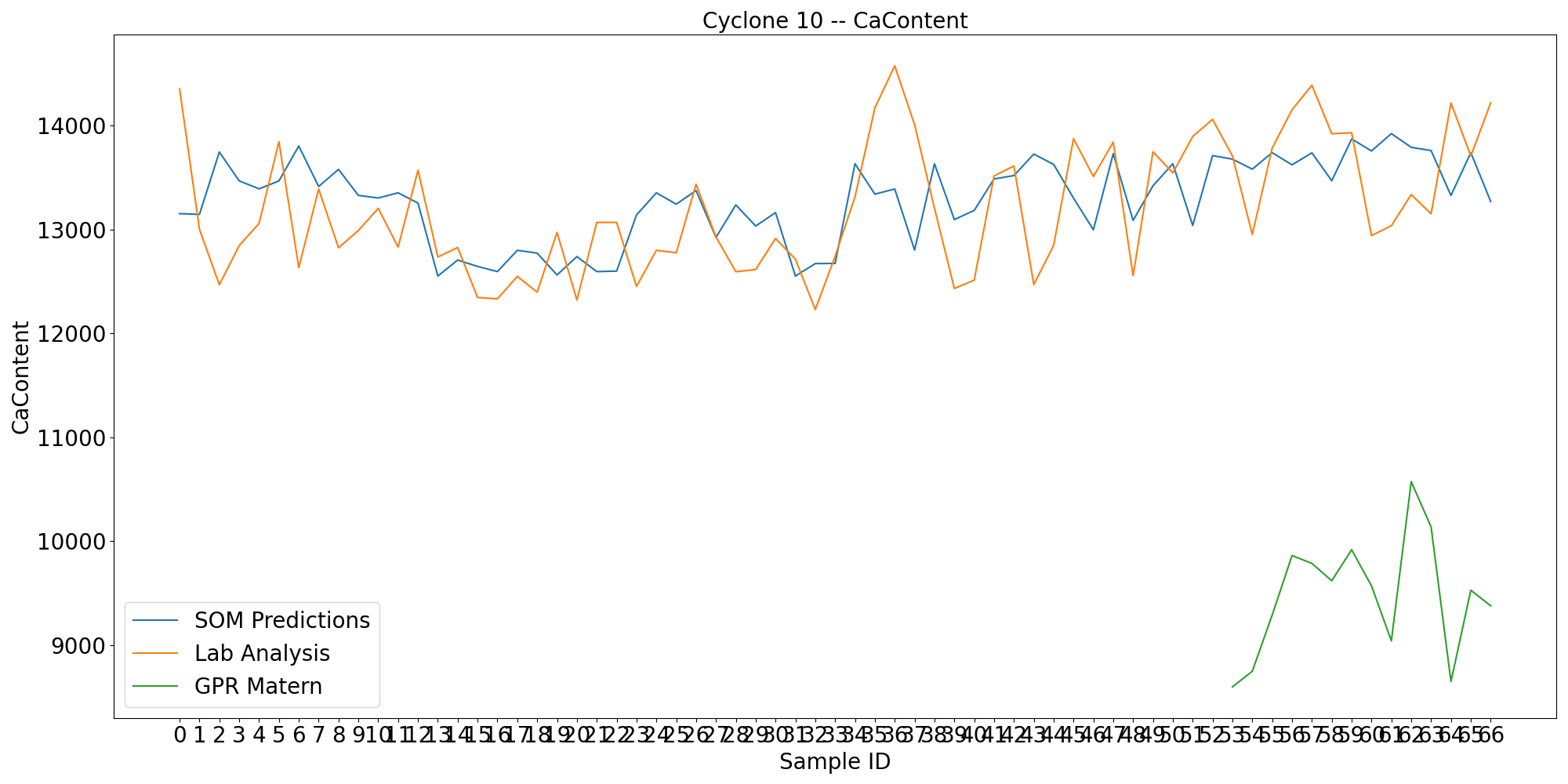}
    \caption{Ca Content}
    \label{subfig:plot5}
  \end{subfigure}
  \hfill
  \begin{subfigure}[b]{0.45\textwidth}
    \centering
    \includegraphics[width=\textwidth]{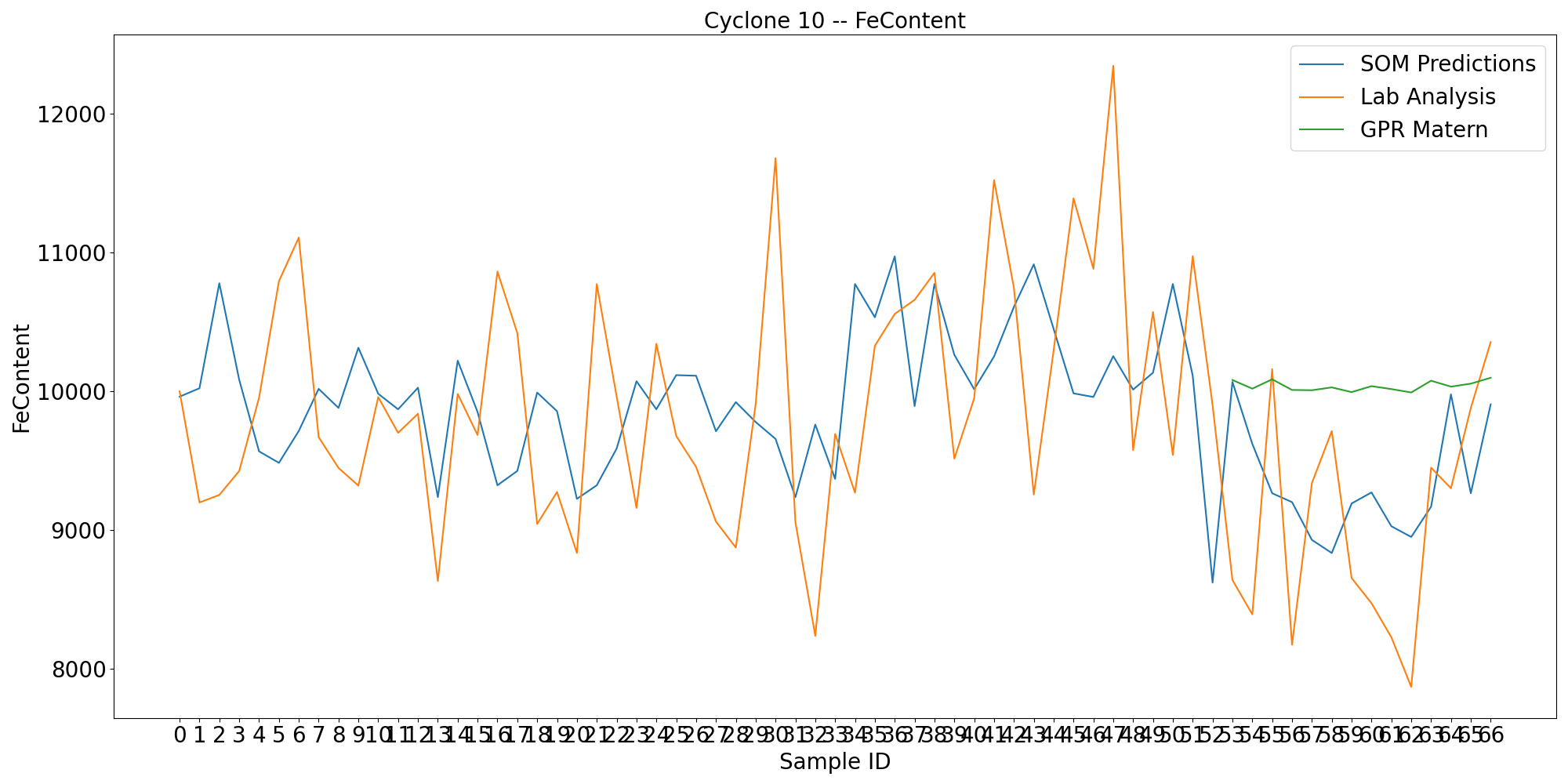}
    \caption{Fe Content}
    \label{subfig:plot6}
  \end{subfigure}

    \vspace{0.5cm}
  
  \begin{subfigure}[b]{0.45\textwidth}
    \centering
    \includegraphics[width=\textwidth]{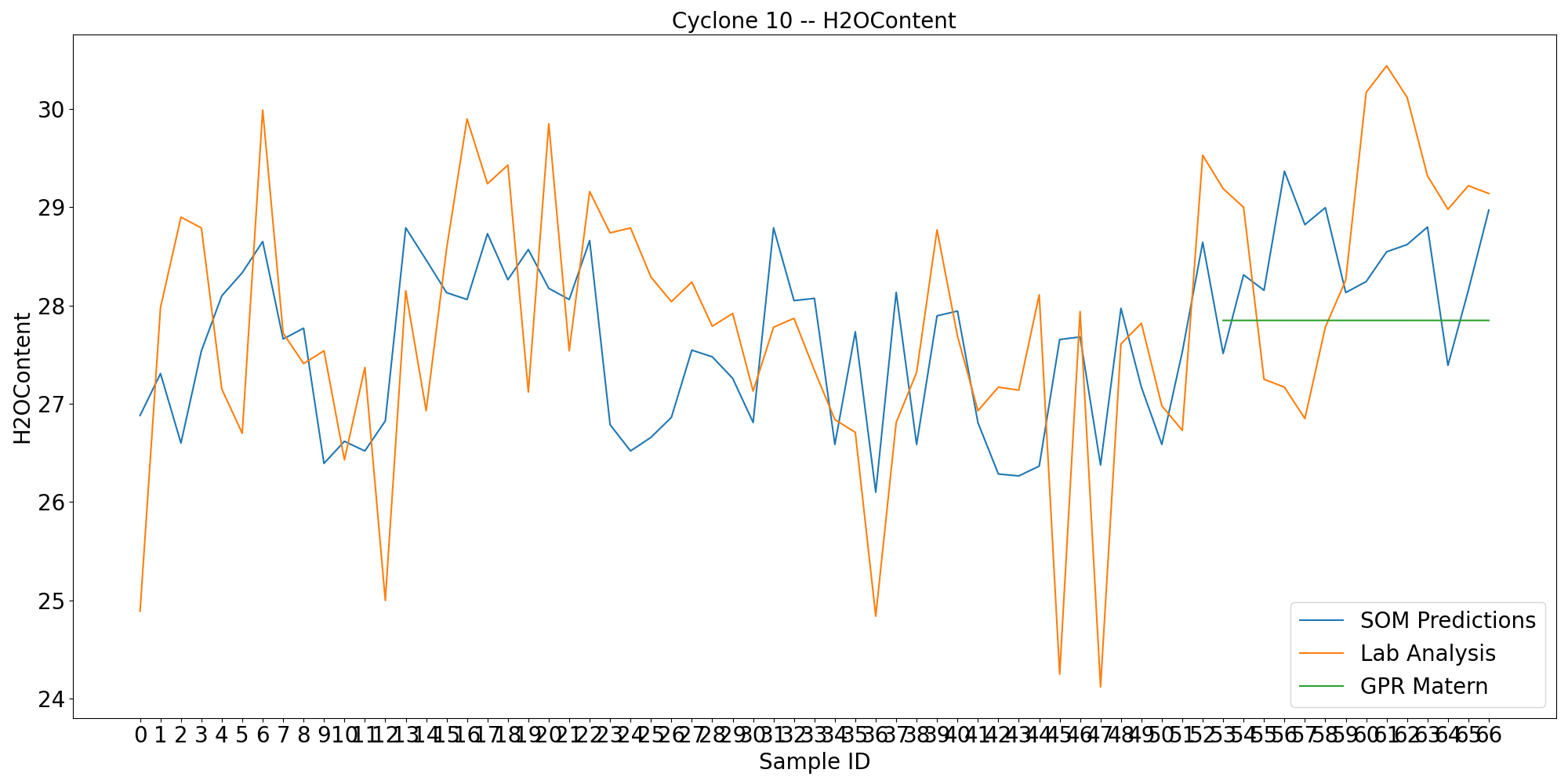}
    \caption{H2O Content}
    \label{subfig:plot7}
  \end{subfigure}
  \hfill
  \begin{subfigure}[b]{0.45\textwidth}
    \centering
    \includegraphics[width=\textwidth]{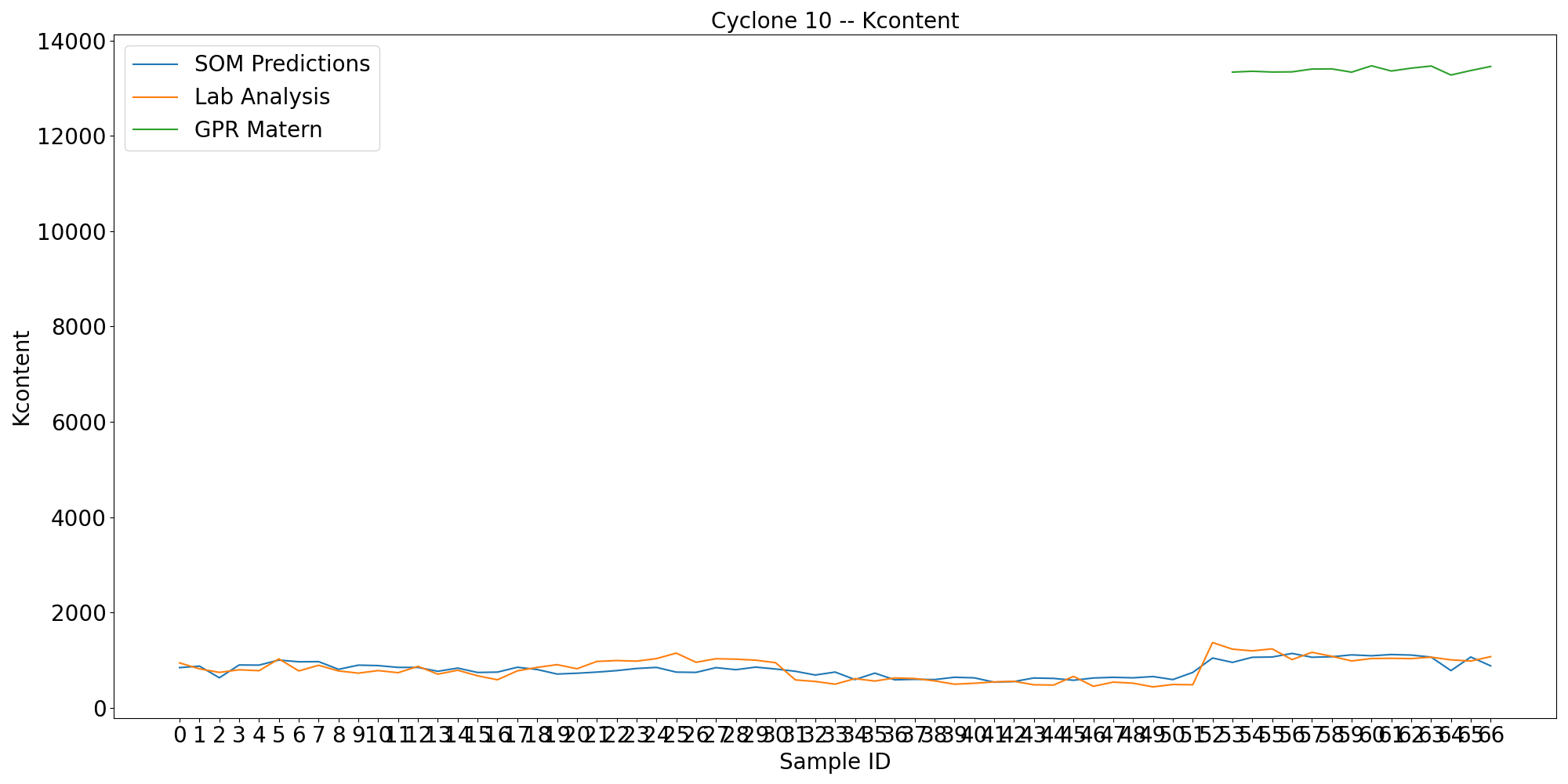}
    \caption{K Content}
    \label{subfig:plot8}
  \end{subfigure}

  \caption{SOM vs GPR Estimated Coal properties}
  \label{fig:fourplots}
\end{figure*}

\begin{figure*}[htbp]
  \centering
  \begin{subfigure}[b]{0.45\textwidth}
    \centering
    \includegraphics[width=\textwidth]{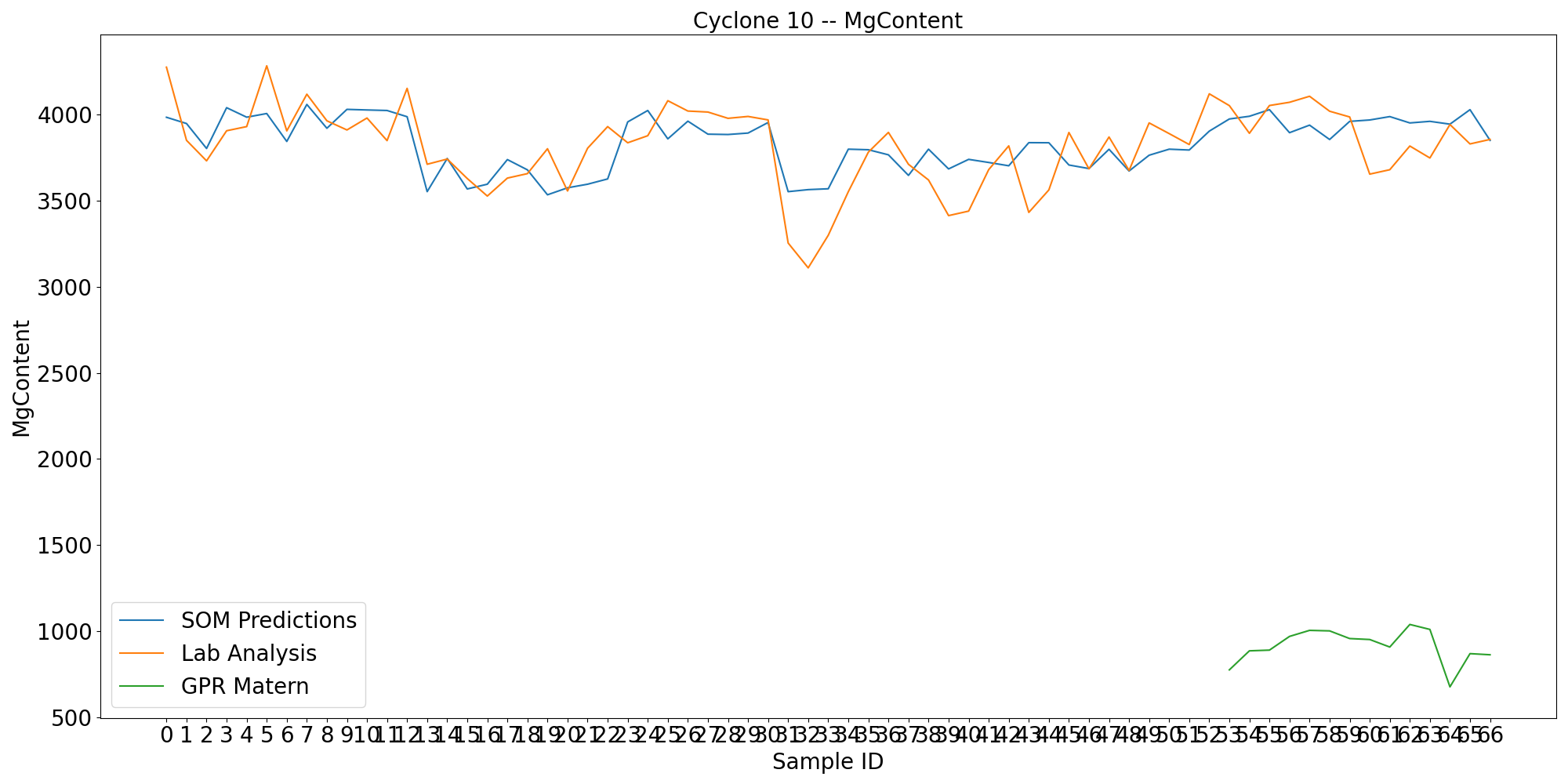}
    \caption{Mg Content}
    \label{subfig:plot1}
  \end{subfigure}
  \hfill
  \begin{subfigure}[b]{0.45\textwidth}
    \centering
    \includegraphics[width=\textwidth]{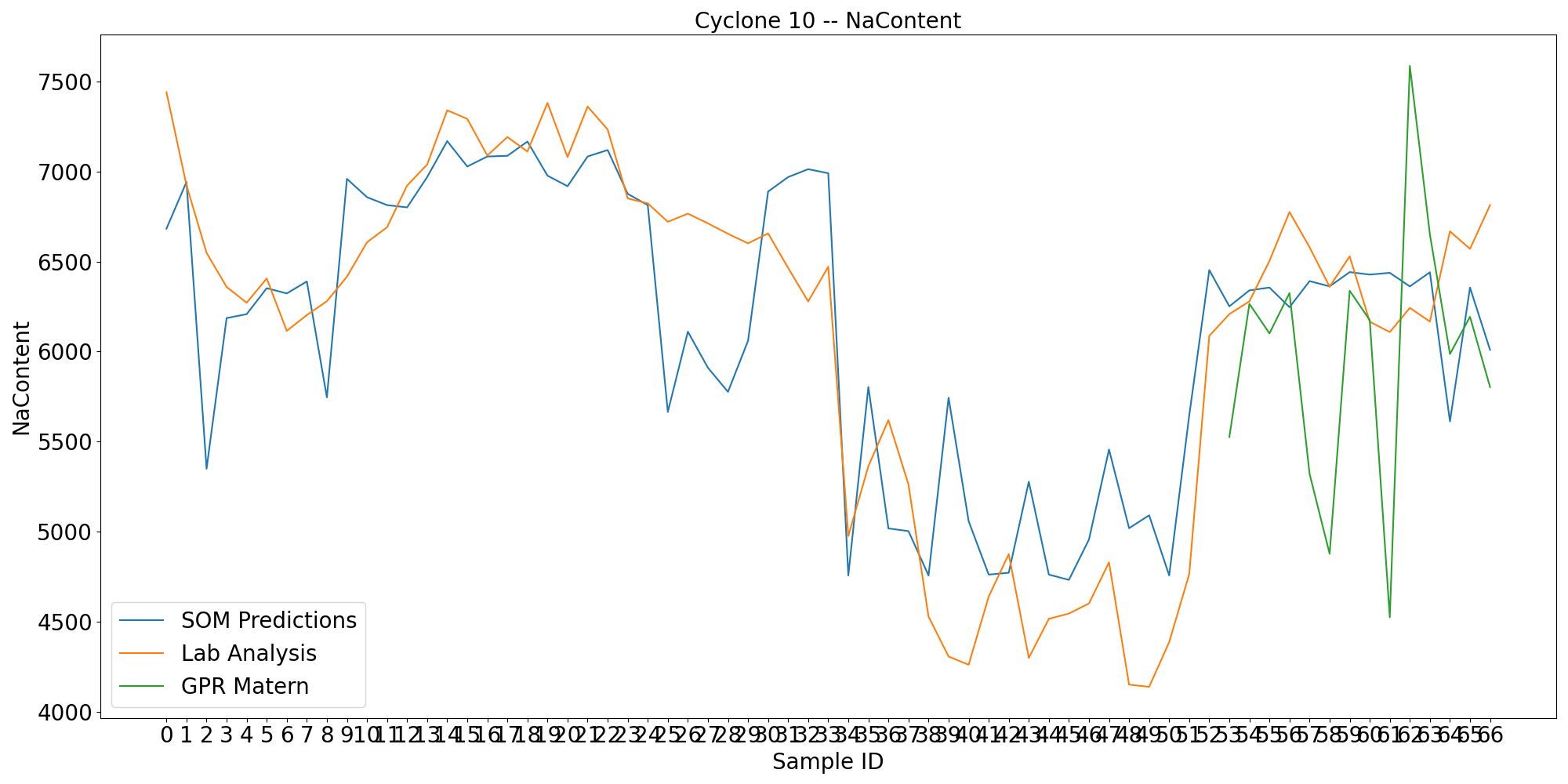}
    \caption{Na Content}
    \label{subfig:plot2}
  \end{subfigure}
  
  \vspace{0.5cm}
  
  \begin{subfigure}[b]{0.45\textwidth}
    \centering
    \includegraphics[width=\textwidth]{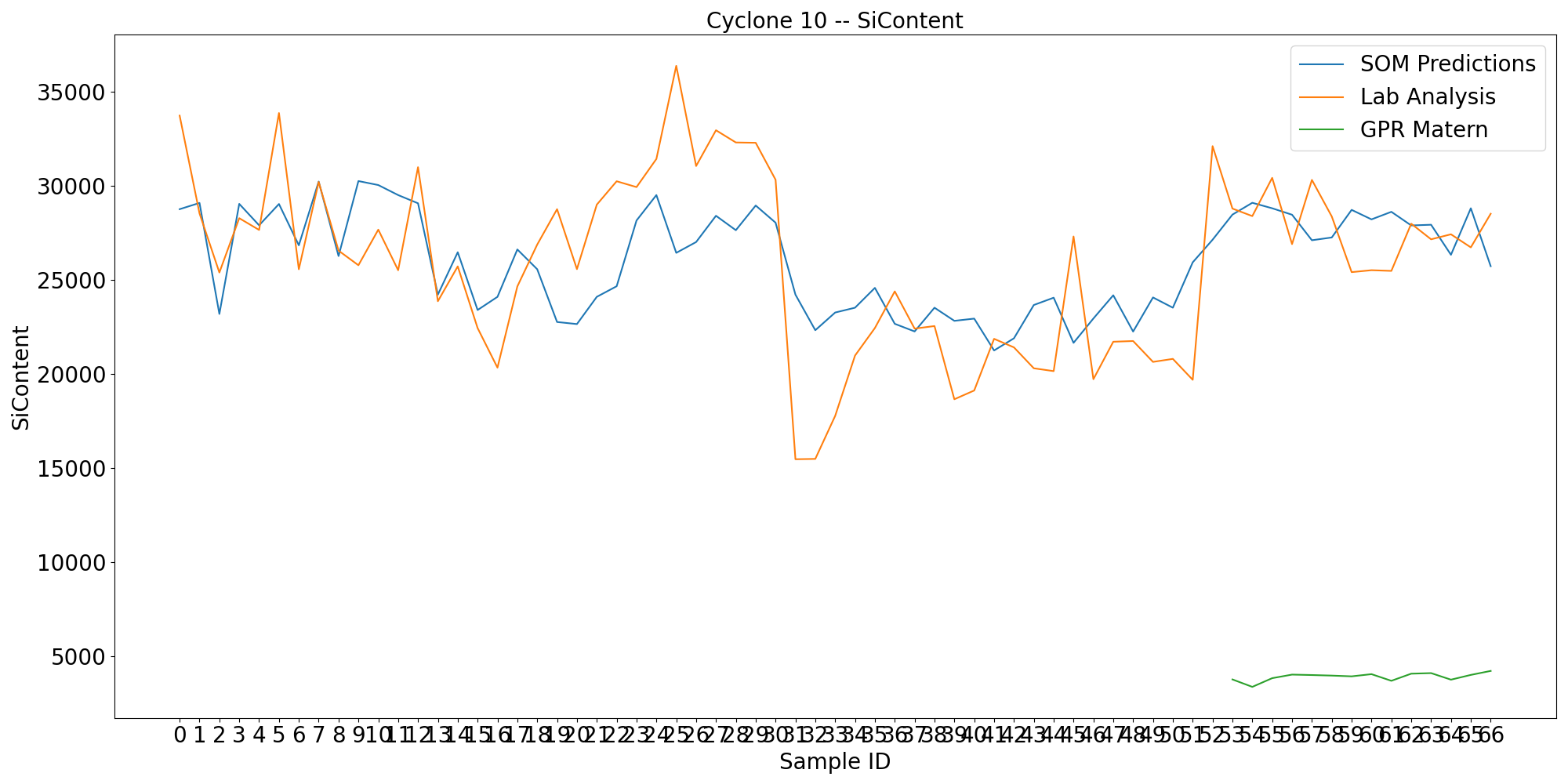}
    \caption{Si Content}
    \label{subfig:plot3}
  \end{subfigure}
  \hfill
  \begin{subfigure}[b]{0.45\textwidth}
    \centering
    \includegraphics[width=\textwidth]{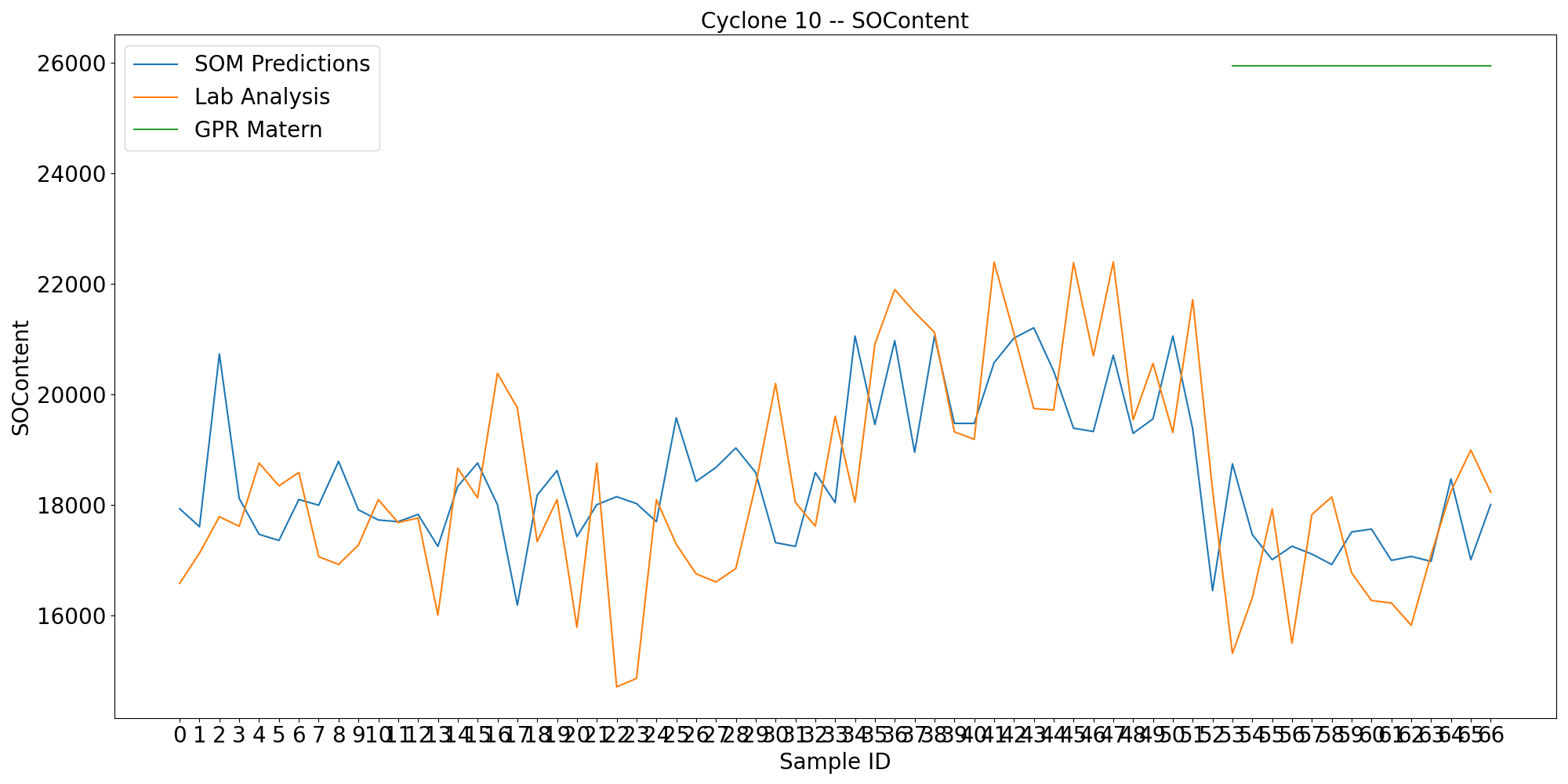}
    \caption{SO Content}
    \label{subfig:plot4}
  \end{subfigure}

  \vspace{0.5cm}
  
  \begin{subfigure}[b]{0.45\textwidth}
    \centering
    \includegraphics[width=\textwidth]{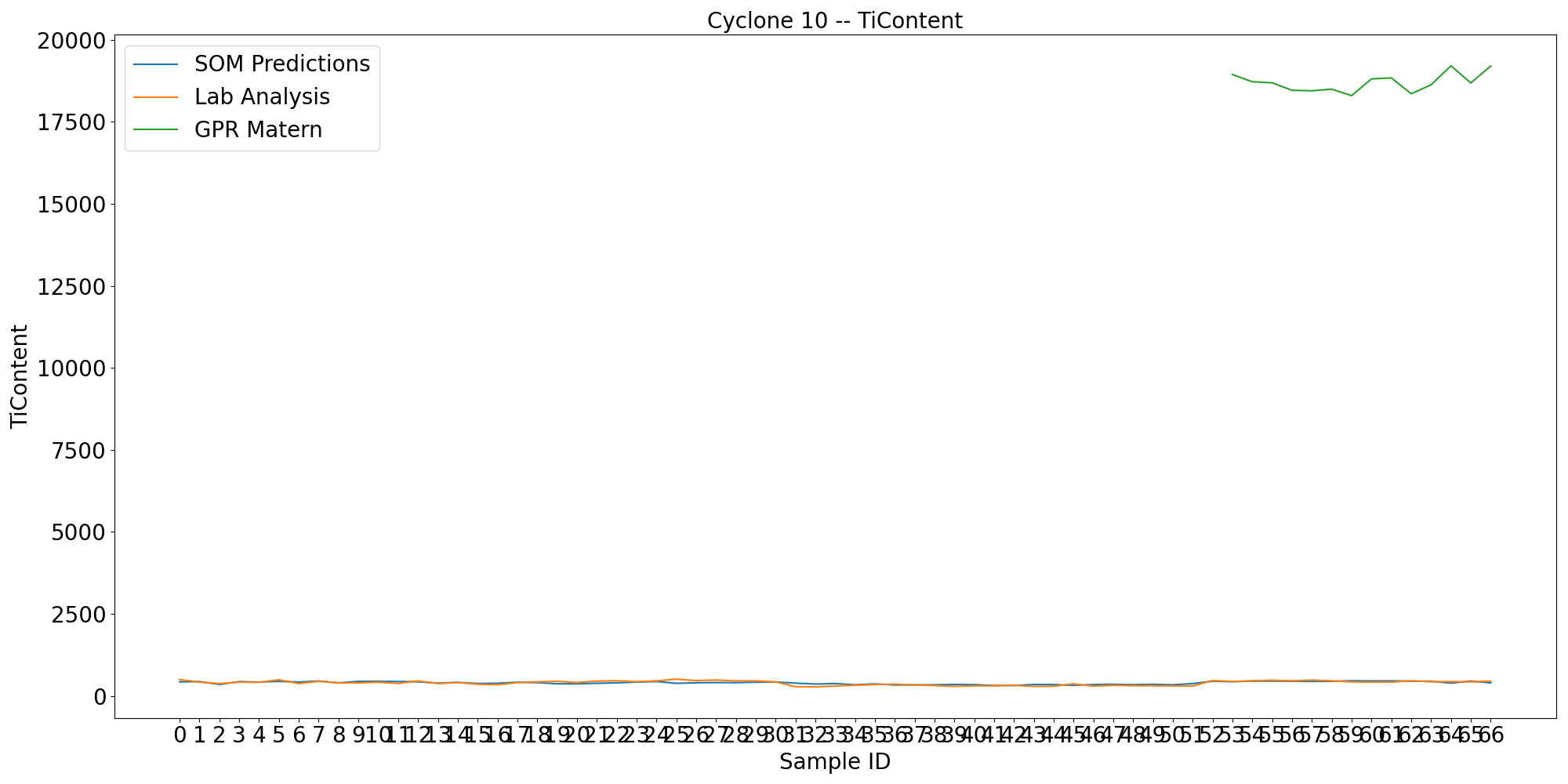}
    \caption{Ti Content}
    \label{subfig:plot5}
  \end{subfigure}
  \hfill

  \caption{SOM vs GPR Estimated Coal properties (Continued)}
  \label{fig:fourplots2}
\end{figure*}

Figure~\ref{fig:fourplots} and~\ref{fig:fourplots2} compare all 13 SOM and GPR estimated coal properties. In the plots, the lab analysis values are the true labels. The SOM results use leave-one-out value estimation, so it is possible to make estimations on all 67 data points. GPR (and the other regression methods), however, use the first 53 datapoints for training. The plots show the validation performance on the 14 testing data points. From the plots we see the GPR performance is not stable on different coal properties. It can make valid estimations for more stable coal properties such as Ash content, however it does not make valid predictions for coal properties such as Al and Ca content. Further, it performs extremely poorly for Na content which has wide oscillations.

\subsection{Topological Regression vs Regression Methods}
Table~\ref{tab:standard_rmse} and \ref{tab:robust_rmse} show the validation prediction RMSE in the original scale of each coal (data) property, across all the experimental methodologies, using Standard and Robust data normalization methods.

\begin{table*}[h]
\centering
\footnotesize
\caption{Standard RMSE Table}
\label{tab:standard_rmse}
\begin{tabular}{c|*{10}{|c}}
\hline
& \multirow{2}{*}{Linear} & \multirow{2}{*}{Polynomial} & \multicolumn{2}{c|}{GPR} & \multirow{2}{*}{DNN} & \multirow{2}{*}{KNN} & \multicolumn{4}{c}{SOM} \\
\cline{4-5}
\cline{8-11}
&  &  & RBF & Matern &  &  & RAND &  AVG &  LS &  WAVG \\
\hline
B/A             & 0.09 & 0.04 & \bf{0.01} & 0.02 & 0.04 & \bf{0.01} & 44.79 & 6.25 & 1.64 & 0.10 \\
Ash             & 0.47 & 0.63 & 0.20 & \bf{0.19} & 0.56 & 0.22 & 4.48e2 & 41.04 & 9.71 & 0.57 \\
BTU             & 9.86e4 & 4.80e4 & 2.92e4 & 2.78e4 & 4.82e4 & 1.18e4 & 1.16e5 & 2.81e4 & 1.89e3 & \bf{1.41e2} \\
H2O             & 6.01 & 4.04 & 2.10 & 2.08 & 2.92 & \bf{1.14} & 7.55e2 & 73.02 & 14.42 & 1.30 \\
Na              & 2.69e6 & 2.64e6 & 1.43e5 & 4.91e5 & 1.90e6 & 1.48e5 & 9.19e5 & 1.49e4 & 9.49e3 & \bf{5.50e2} \\
Fe              & 3.30e6 & 1.28e6 & 1.13e6 & 1.38e6 & 2.70e6 & 1.00e6 & 1.10e6 & 7.27e4 & 3.16e4 & \bf{8.50e2} \\
Al              & 3.23e6 & 1.23e7 & 7.01e5 & 6.63e5 & 3.16e6 & 4.58e5 & 2.12e6 & 1.27e5 & 1.60e4 & \bf{1.03e3} \\
Ca              & 7.83e5 & 1.53e6 & 4.55e5 & 3.64e5 & 7.62e5 & 1.99e5 & 9.21e5 & 9.46e4 & 1.91e4 & \bf{5.63e2} \\
K               & 1.42e5 & 5.36e5 & 7.43e4 & 4.88e4 & 1.08e5 & 2.96e4 & 3.25e5 & 5.07e4 & 2.62e3 & \bf{1.54e2} \\
Mg              & 1.20e5 & 7.69e4 & 2.86e4 & 4.51e4 & 6.51e4 & 2.52e4 & 3.73e5 & 7.08e4 & 6.30e3 & \bf{1.91e2} \\
Si              & 2.19e7 & 1.12e8 & 5.44e6 & 5.19e6 & 3.33e7 & 4.95e6 & 9.32e6 & 2.01e6 & 4.20e4 & \bf{3.57e3} \\
SO              & 1.52e7 & 3.59e6 & 3.00e6 & 3.17e6 & 1.08e7 & 2.24e6 & 3.82e6 & 1.12e6 & 1.77e4 & \bf{1.56e3} \\
Ti              & 1.11e4 & 4.21e4 & 2.91e3 & 3.46e3 & 5.49e3 & 5.65e2 & 1.36e5 & 3.57e4 & 6.27e2 & \bf{46.65} \\
\hline
Average         & 3.65e6 & 1.03e7 & 8.46e5 & 8.75e5 & 4.06e6 & 6.97e5 & 1.47e6 & 2.79e5 & 1.13e4 & \bf{6.66e2} \\
\hline
\end{tabular}
\end{table*}

\begin{table*}[h]
\centering
\footnotesize
\caption{Robust RMSE Table}
\label{tab:robust_rmse}
\begin{tabular}{c|*{10}{|c}}
\hline
& \multirow{2}{*}{Linear} & \multirow{2}{*}{Polynomial} & \multicolumn{2}{c|}{GPR} & \multirow{2}{*}{DNN} & \multirow{2}{*}{KNN} & \multicolumn{4}{c}{SOM} \\
\cline{4-5}
\cline{8-11}
&  &  & RBF & Matern &  &  & RAND &  AVG &  LS &  WAVG \\
\hline
B/A        & 0.08   & 0.15       & \bf{0.01}   & 0.02       & 0.05 & \bf{0.01} & 26.62    & 25.71   & 0.34   & 0.12 \\
Ash        & 0.64   & 0.42       & 0.21      & 0.20      & 0.52 & \bf{0.19} & 2.80e2   & 1.43e2  & 1.30   & 0.59 \\
BTU        & 8.05e4 & 4.02e4     & 2.82e4    & 2.80e4     & 1.21e5 & 1.19e4 & 9.35e4   & 3.78e4  & 4.50e2 & \bf{1.61e2} \\
H2O        & 4.98   & 3.32       & 2.21      & 2.19       & 4.99 & \bf{1.04} & 5.01e2   & 2.20e2  & 3.59   & 1.38 \\
Na         & 2.65e6 & 4.67e6     & 5.26e4    & 2.64e5     & 2.68e6 & 1.68e5 & 6.27e5   & 3.44e5  & 1.95e3 & \bf{6.10e2} \\
Fe         & 2.52e6 & 1.23e6     & 1.01e6    & 1.51e6     & 3.58e6 & 1.16e6 & 6.95e5   & 3.37e5  & 2.11e3 & \bf{8.41e2} \\
Al         & 3.75e6 & 5.01e6     & 4.54e5    & 4.40e5     & 2.86e6 & 4.58e5 & 1.88e6   & 7.74e5  & 2.92e3 & \bf{1.08e3} \\
Ca         & 1.00e6 & 1.33e6     & 6.17e5    & 5.13e5     & 1.09e6 & 2.21e5 & 7.73e5   & 3.51e5  & 1.39e3 & \bf{5.85e2} \\
K          & 1.40e5 & 5.37e5     & 8.74e4    & 4.94e4     & 1.36e5 & 2.73e4 & 3.29e5   & 1.32e5  & 4.21e2 & \bf{1.56e2} \\
Mg         & 1.17e5 & 7.31e4     & 2.47e4    & 3.40e4     & 7.13e4 & 2.50e5 & 2.66e5   & 1.16e5  & 4.40e2 & \bf{2.01e2} \\
Si         & 2.88e7 & 5.59e7     & 3.68e6    & 5.66e6     & 3.78e7 & 3.49e6 & 7.35e6   & 3.13e6  & 9.27e3 & \bf{3.69e3} \\
SO         & 1.11e7 & 3.19e6     & 2.39e6    & 2.53e6     & 1.05e7 & 2.22e6 & 2.77e6   & 1.15e6  & 4.48e3 & \bf{1.50e3} \\
Ti         & 1.31e4 & 1.95e4     & 1.24e3    & 1.45e3     & 8.81e3 & 7.80e2 & 1.30e5   & 5.71e4  & 1.30e2 & \bf{48.23} \\
\hline
Average    & 3.85e6 & 5.54e6     & 6.42e5    & 8.49e5     & 4.52e6 & 5.99e5 & 1.15e6   & 4.95e5  & 1.81e3 & \bf{6.83e2} \\
\hline
\end{tabular}
\end{table*}

\subsection{DBSCAN}

\begin{figure}[!t]
  \centering
  \includegraphics[width=\columnwidth]{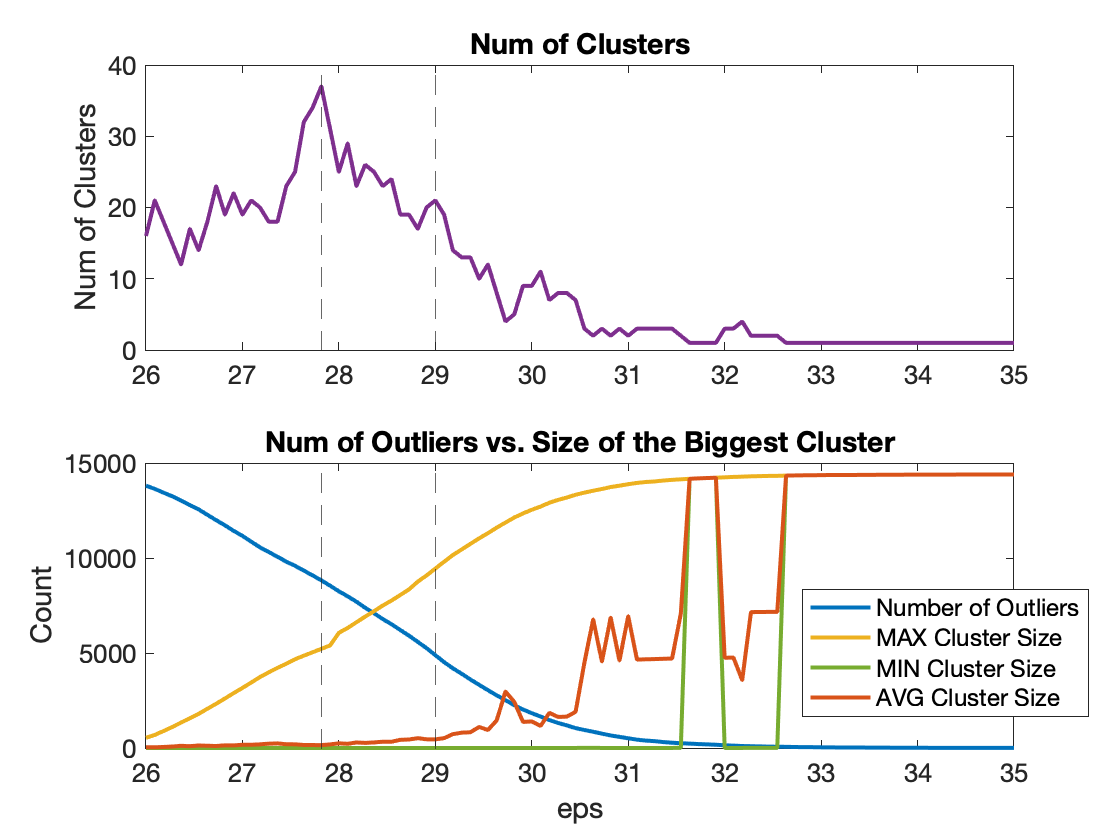}
  \caption{DBSCAN without PCA.}
  \label{fig:dbscan}
\end{figure}

Figure \ref{fig:dbscan} show the relationship of using different $eps$ and the number of clusters, size of the biggest, average, and smallest cluster, and the number of outliers on the original data without utilizing PCA.

\end{document}